\documentclass[10pt,journal,compsoc]{IEEEtran}
\ifCLASSOPTIONcompsoc
  \usepackage[nocompress]{cite}
\else
  \usepackage{cite}
\fi

\ifCLASSINFOpdf
  \usepackage[pdftex]{graphicx}
\else
\fi
\usepackage{hyperref}
\usepackage{amsmath}
\usepackage{amssymb}
\usepackage{amsthm}
\usepackage{booktabs}
\usepackage{bm} 
\usepackage{multirow}
\usepackage{colortbl}

\hypersetup{hidelinks} 

\DeclareMathOperator*{\argmax}{arg\,max}

\usepackage[linesnumbered,ruled,vlined,norelsize]{algorithm2e}
\makeatletter
\newcommand{\removelatexerror}{\let\@latex@error\@gobble}
\makeatother

\SetCommentSty{mycommfont}

\SetKwInput{KwInput}{Input}                
\SetKwInput{KwOutput}{Output}              
\SetKwInput{KwNotations}{Notations}        

\usepackage{array}
\usepackage{colortbl}

\ifCLASSOPTIONcompsoc
 \usepackage[caption=false,font=footnotesize,labelfont=sf,textfont=sf]{subfig}
\else
 \usepackage[caption=false,font=footnotesize]{subfig}
\fi

\begin{document}
%
\title{DeViT: Decomposing Vision Transformers for Collaborative Inference in Edge Devices}

%
%
%
%

\author{Guanyu Xu, Zhiwei Hao,
        Yong Luo,~\IEEEmembership{Member,~IEEE,}
        Han Hu,~\IEEEmembership{Member,~IEEE,}
        Jianping An,~\IEEEmembership{Member,~IEEE,}
        and~Shiwen Mao,~\IEEEmembership{Fellow,~IEEE}
\IEEEcompsocitemizethanks{
  \IEEEcompsocthanksitem Guanyu Xu, Zhiwei Hao, Han Hu and Jianping An are with the School of Information and Electrionics, Beijing Institute of Technology, Beijing 100081, China. 
  E-mail: \{xuguanyu, haozhw, hhu, an\}@bit.edu.cn.
  \IEEEcompsocthanksitem Yong Luo is with School of Computer Science, Wuhan University, Wuhan 430072, China. 
  E-mail: yluo180@gmail.com.
  \IEEEcompsocthanksitem Shiwen Mao is with Department of Electrical and Computer Engineering, Auburn University, Auburn, AL 36849-5201, USA.
  E-mail: smao@ieee.org.}
\thanks{Manuscript received ...}}

%
%

\markboth{IEEE Transactions on Mobile Computing}%
{Xu \MakeLowercase{\textit{et al.}}: DeViT: Decomposing Vision Transformers for Collaborative Inference in Edge Devices }
%



\IEEEtitleabstractindextext{%
\begin{abstract}
  Recent years have witnessed the great success of vision transformer (ViT), which has achieved state-of-the-art performance on multiple computer vision benchmarks.
  However, ViT models suffer from vast amounts of parameters and high computation cost, leading to difficult deployment on resource-constrained edge devices. 
  Existing solutions mostly compress ViT models to a compact model but still cannot achieve real-time inference. 
  To tackle this issue, we propose to explore the divisibility of transformer structure, 
  and decompose the large ViT into multiple small models for collaborative inference at edge devices. 
  Our objective is to achieve fast and energy-efficient collaborative inference while maintaining comparable accuracy compared with large ViTs.
  To this end, we first propose a collaborative inference framework termed \textbf{DeViT} to facilitate edge deployment by decomposing large ViTs.
  Subsequently, we design a decomposition-and-ensemble algorithm based on knowledge distillation, termed DEKD, to fuse multiple small decomposed models while dramatically reducing communication overheads, 
  and handle heterogeneous models by developing a feature matching module to promote the imitations of decomposed models from the large ViT.
  Extensive experiments for three representative ViT backbones on four widely-used datasets demonstrate our method achieves efficient collaborative inference for ViTs
  and outperforms existing lightweight ViTs, striking a good trade-off between efficiency and accuracy. 
  For example, our DeViTs improves end-to-end latency by 2.89$\times$ with only 1.65\% accuracy sacrifice using CIFAR-100 compared to the large ViT, ViT-L/16, on the GPU server. 
  DeDeiTs surpasses the recent efficient ViT, MobileViT-S, by 3.54\% in accuracy on ImageNet-1K, 
  while running 1.72$\times$ faster and requiring 55.28\% lower energy consumption on the edge device.
\end{abstract}

\begin{IEEEkeywords}
 Vision transformer, model decomposition, edge computing, collaborative inference
\end{IEEEkeywords}}

\maketitle

\IEEEdisplaynontitleabstractindextext

%
\IEEEpeerreviewmaketitle

\IEEEraisesectionheading{\section{Introduction}\label{sec:introduction}}

\IEEEPARstart{T}{ransformers}\cite{transformer} have been widely used in natural language processing (NLP) \cite{bert}, achieved exceptional performance in audio tasks\cite{ast}, 
and demonstrated their great potentials in computer vision\cite{swin_transformer,vit}.
With the rapid development of the artificial intelligence of things (AIoT), deploying transformer-based models on resource-limited edge devices has becoming an attractive idea. 
However, the inference cost of transformer is very high, leading to unacceptable latency or energy consumption, especially for resource-limited edge devices\cite{DBLP:journals/tmc/MasoudiC21}.
For example, the ViT-L/16 model \cite{vit}, a typical vision transformer, requires 1.12 GB storage and 190.7 Giga Floating Point Operations (GFLOPs) for inference, 
while a Raspberry Pi-4B device only has 4 GB RAM and 13.5 Giga Floating Point Operations Per Second (GFLOPS) \cite{gpu_devices,pi_device}.
In order to achieve effective inference, a pre-trained model is usually divided into multiple parts, and each part is deployed on an edge device or a server. 
These devices conduct inference collaboratively. This is often called collaborative inference, 
which avoids transmitting large data to the server, 
and can substantially improve the quality of experience in the AIoT\cite{Neurosurgeon}.

In collaborative inference, some methods partition DNNs by or across layers\cite{Neurosurgeon, DBLP:conf/icml/KimPKH17}. 
The intermediate features should be transmitted to devices or servers multiple times, 
resulting in high latency and energy costs.
The data transmission can also be easily affected by the network dynamics. 
Furthermore, some complex surgery strategies are usually required to find a proper point to decouple the original model\cite{DBLP:conf/infocom/MohammedJBF20}.
Therefore, achieving fast and effective collaborative inference on edge devices is challenging. 

Vision transformers (ViTs) cannot be directly deployed on edge devices because of the large model size and high computation cost. 
To tackle this problem, 
some existing works utilize conventional model compression methods to assist the deployment of large ViTs and achieve fast inference. 
These methods need to elaborately design\cite{mobile_vit} or search\cite{autoformer} for the compact structure, 
and cannot balance between accuracy and the constrained resource requirements of edge devices. 
This is because these methods focus on reducing model parameters and FLOPs, 
which yet are not the right criterion for the ultimate goals of latency and energy. 
Moreover, almost no on-device experiments have been conducted so far to measure the actual inference latency and energy consumption.  

To remedy these limitations, we propose a collaborative inference framework for general ViTs in edge devices, termed DeViT, by decomposing the large ViT into multiple small models. 
These decomposed models can be readily deployed on edge devices for conducting inference simultaneously, which significantly reduces inference latency and energy consumption with slight accuracy drop. 
To the best of our knowledge, this is the first work to achieve collaborative inference for ViTs in edge devices by leveraging the divisibility of the transformer structure. 
Subsequently, we propose an efficient decomposition-and-ensemble algorithm based on knowledge distillation, termed DEKD, 
to counterbalance the accuracy loss induced by decomposition and facilitate communication overheads among edge devices. 
Each device equipped with a decomposed model only needs to execute a singular transfer of target features for aggregation, resulting in a reduction of transmission costs alongside a satisfactory ensemble performance. 
To ensure acceptable performance, we also design a novel feature matching module to mitigate information in DEKD algorithm, 
which promotes the imitations of decomposed models from the large transformer by matching the dimension of intermediate features between them.

We conduct comprehensive experiments on both GPU servers and edge devices to verify the effectiveness of our proposed framework. 
In the experiments, we first decompose a large ViT according to the number of edge devices and their hardware configuration. 
Then we utilize DEKD to retain the most important parts in the small models by shrinking and finally fuse these small models by employing our feature aggregation module. 
The results show that the proposed DeViT framework can achieve real-time inference and almost no accuracy drop on edge devices for large ViTs. 
Moreover, we conduct massive experiments using three ViT backbones and four computer vision datasets to verify the effectiveness of our approach. 
For instance, on the GPU server, our DeViTs improves end-to-end latency by 2.89$\times$ with only 1.65\% accuracy sacrifice compared to the large ViT, ViT-L/16 \cite{vit}, using ImageNet. 
On the edge device, DeDeiTs surpasses the recent efficient ViT, MobileViT-S \cite{mobile_vit}, by 3.54\% in accuracy, while running 1.72$\times$ faster and requiring 55.28\% lower energy consumption.

Our main contributions in this paper are summarized as follows: 
\begin{itemize}
  \item We propose a collaborative inference framework for general ViTs in edge devices, termed DeViT, by decomposing the large ViT into multiple small models.
  \item We develop an efficient decomposition-and-ensemble algorithm based on knowledge distillation, termed DEKD, to fuse multiple decomposed models and minimize the accuracy loss caused by decomposition, 
  while dramatically reducing communication overheads among edge devices. 
  \item We design a novel feature matching module that facilitates the learning of decomposed models from the large ViT and minimizes information loss. 
  \item We evaluate our DeViT framework for three representative ViT backbones on both commodity GPU servers and real edge devices using four widely-used datasets. 
  The results demonstrate that our method can reduce inference latency by $1.90 \times$ and reduce energy consumption by 26.11\% on average with only 2 \% accuracy sacrifice. 
\end{itemize}

The remainder of this paper is organized as follows. 
We elaborate the challenges for edge deployment of ViTs in Section \ref{sec:challenges}. 
The proposed collaborative deployment framework of ViTs on multiple edge devices is presented in Section \ref{sec:system}. 
The proposed DEKD algorithm is described in Section \ref{sec:algorithm}, 
and then Section \ref{sec:experiment} provides our experiments and analysis. 
Finally, we discuss the related works in Section \ref{sec:related_works} and conclude our paper in Section \ref{sec:conclusion}.

\section{Challenges of Edge Deployment for ViTs}
\label{sec:challenges}

In this section, we analyze the number of parameters for ViTs, 
and discuss the challenges of edge deployment.

\begin{table*}
  \renewcommand\arraystretch{1.2}
  \centering
  \caption{Comparison Between Resources on Typical Edge Devices and Deployment Requirements for Representative ViTs}
  \label{tab:table1}
  \begin{tabular}{c|ccc||c|ccc} 
  \toprule
  \multicolumn{4}{c||}{\textbf{Resources on Typical Edge Devices}}                                            & \multicolumn{4}{c}{\begin{tabular}[c]{@{}c@{}}\textbf{Required Computation Consumption for Deploying}\\ \textbf{Representative ViTs~}\end{tabular}}  \\ 
  \hline
  Devices                 & Memory & \begin{tabular}[c]{@{}c@{}}Computation\\Performance\end{tabular} & TDP   & Models   & \#Parameter & Storage & FLOPs                                                                                                             \\ 
  \hline
  Raspberry Pi-4B \cite{pi_device}         & 4 GB   & 13.5 GFLOPS                                                      & 7.3 W & DeiT-B \cite{deit}   & 86 M        & 333 MB  & 55.4 G                                                                                                            \\
  NVIDIA Jetson Nano \cite{gpu_devices}      & 4 GB   & 472 GFLOPS                                                       & 10 W  & Swin-B \cite{swin_transformer}   & 88 M        & 348 MB  & 47.0 G                                                                                                            \\
  NVIDIA Jetson TX2 \cite{gpu_devices}       & 8 GB   & 1.33 TFLOPS                                                      & 15 W  & Swin-L \cite{swin_transformer}   & 197 M       & 763 MB  & 103.9 G                                                                                                           \\
  NVIDIA Jetson Xavier NX \cite{nvidia_xavier_nx} & 16 GB  & 21 TOPS                                                          & 20 W  & ViT-L/16 \cite{vit} & 304 M       & 1.12 GB & 190.7 G                                                                                                           \\
  \bottomrule
  \end{tabular}
  \end{table*}

\begin{figure}
  \centering
  \includegraphics[width=0.475\textwidth]{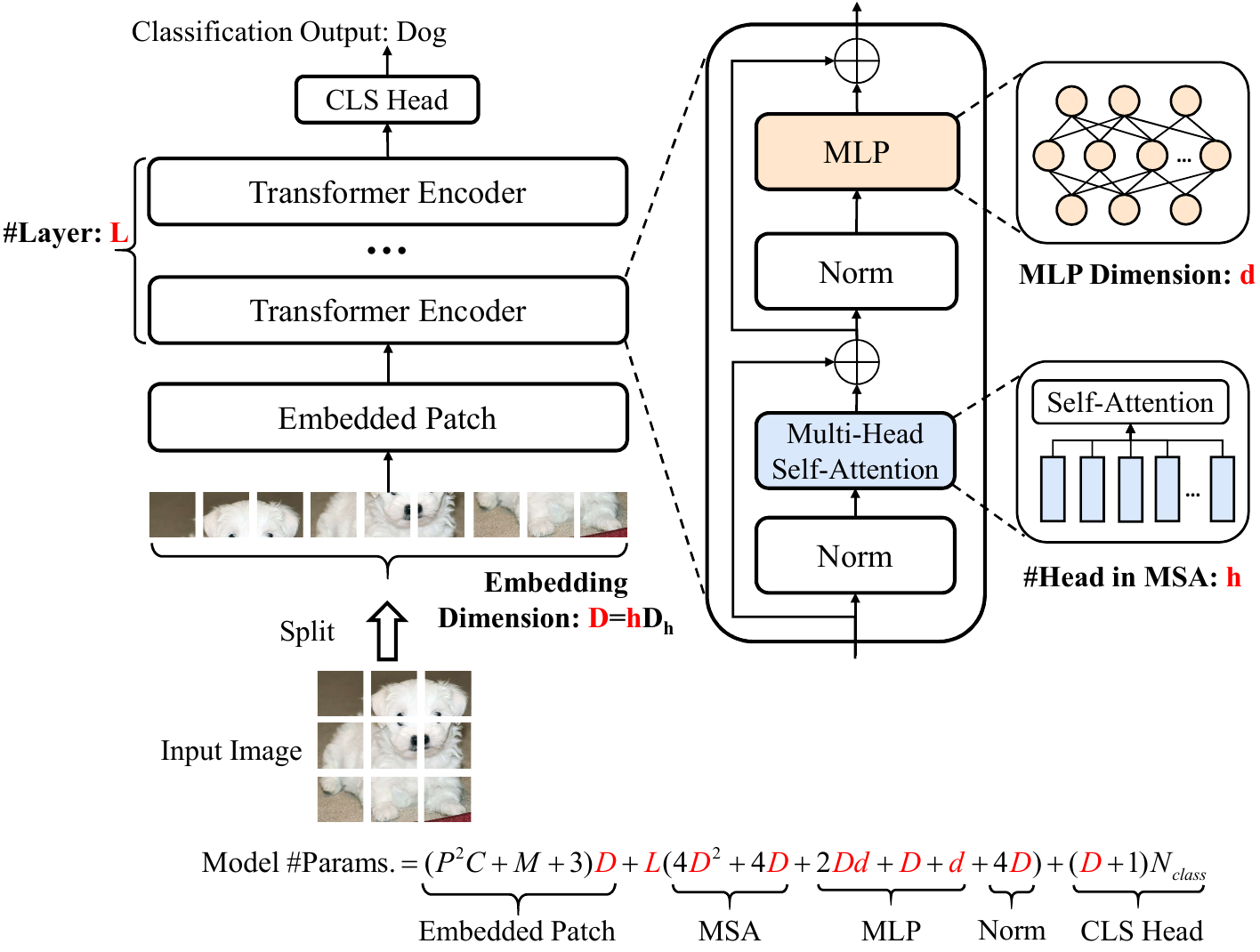}
  \caption{The structure of a ViT\cite{vit}. 
  ViT is composed of a patch embedding layer, several transformer encoders, and a classification head.
  The model size mainly depends on the number of encoders, 
  the embedding dimension, the number of attention heads, and the MLP dimension.}
  \label{fig:vit}
\end{figure}

\subsection{Parameter Analysis of ViTs}

The ViT model consists of a patch embedding layer, multiple stacked transformer encoders, and a classification head, as shown in Fig. \ref{fig:vit}. 
We introduce the structure of ViT in the Appendix \ref{sc:vit_intro}. 
The number of parameters depends on the embedding dimension $D$, the number of transformer encoder $L$, MLP hidden dimension $d$, and the number of heads $h$. 
The parameters of the embedded patch are $(P^2C+M+3)D$, where $P$, $C$ and $M$ is the size of patches, the number of channel and the number of patches, respectively. 
In a transformer encoder, the parameters of MSA and MLP are $4D^2+4D$ and $2Dd+D+d$, respectively. 
The number of parameters of the two LN modules are $4D$. 
Given the classification task, we suppose the number of classes is $N_{class}$. 
To sum up, the total number of parameters for the entire model is 
\begin{equation}
  \begin{split}
    \#\text{Params.} = &(P^2C+M+3)D+\\
    &L(4D^2+4D+2Dd+D+d+4D)+\\
    &(D+1)N_{class},\\
  \end{split}
\end{equation}
where $D=hD_h$. 
The model size mainly depends on the number of network layer $L$, 
the embedding dimension $D$, the number of attention heads $h$, and the MLP dimension $d$.
Taking the ViT-L/16 to perform a 1000-class classification task as an example, 
the size of patches, the number of patches, the number of network layer, the embedding dimension, 
the number of attention heads and the MLP dimension are 16, 196, 24, 1024, 16, and 4096, respectively. 
Thus, the number of ViT-L/16 parameters is 304 M. 
If the parameters of ViT-L/16 are stored as the 32-bit floating point, it requires a storage space of 1.12 GB.

\subsection{Challenges to Edge Deployment}
Although ViTs have achieved the state-of-the-art performance in most vision tasks, 
it is difficult to deploy them on resource-constrained devices.
We compare the resources of typical edge devices with the deployment requirements for representative ViTs in Table \ref{tab:table1}, 
where we observe three major challenges:
\begin{itemize}
  \item \textbf{Limited memory and storage capacity.} 
  Different from GPU servers, edge devices usually have low memory and storage capacity. 
  For example, ViT-L/16 needs at least 2.60 GB memory and 1.12 GB storage space for inference, 
  while the Raspberry Pi-4B device has only 4 GB memory. 
  \item \textbf{Huge computation cost.} 
  The computation cost of ViTs is massive due to their complicated stacked computational structure. 
  For instance, inference with ViT-L/16 requires 190.7 GFLOPs, 
  but the computation capacity of Raspberry Pi-4B  is only 13.5 GFLOPS. 
  \item \textbf{High inference latency and energy cost.} 
  The inference of ViTs brings about high latency and energy cost.
  In order to classify an image with 224$\times$224 resolution, 
  the ViT-L/16 model takes 56.79 milliseconds and 0.58 J on an NVIDIA Jetson TX2 device.
\end{itemize}

In consideration of these challenges, 
we aim to design a collaborative edge deployment framework for ViTs with the following design goals: 
\begin{itemize}
  \item \textbf{Generality. }The framework can be applied to different variants of ViTs. 
  \item \textbf{Lightweight. }
  The framework can accommodate different hardware platforms by reducing the storage and computation cost. 
  \item \textbf{Rapidity. }The framework can meet the latency requirement of IoT applications with a fast inference speed. 
  \item \textbf{Reliability. }The framework can achieve a performance comparable to that of powerful giant ViTs.
\end{itemize}

\begin{figure}
    \centering
    \includegraphics[width=0.48\textwidth]{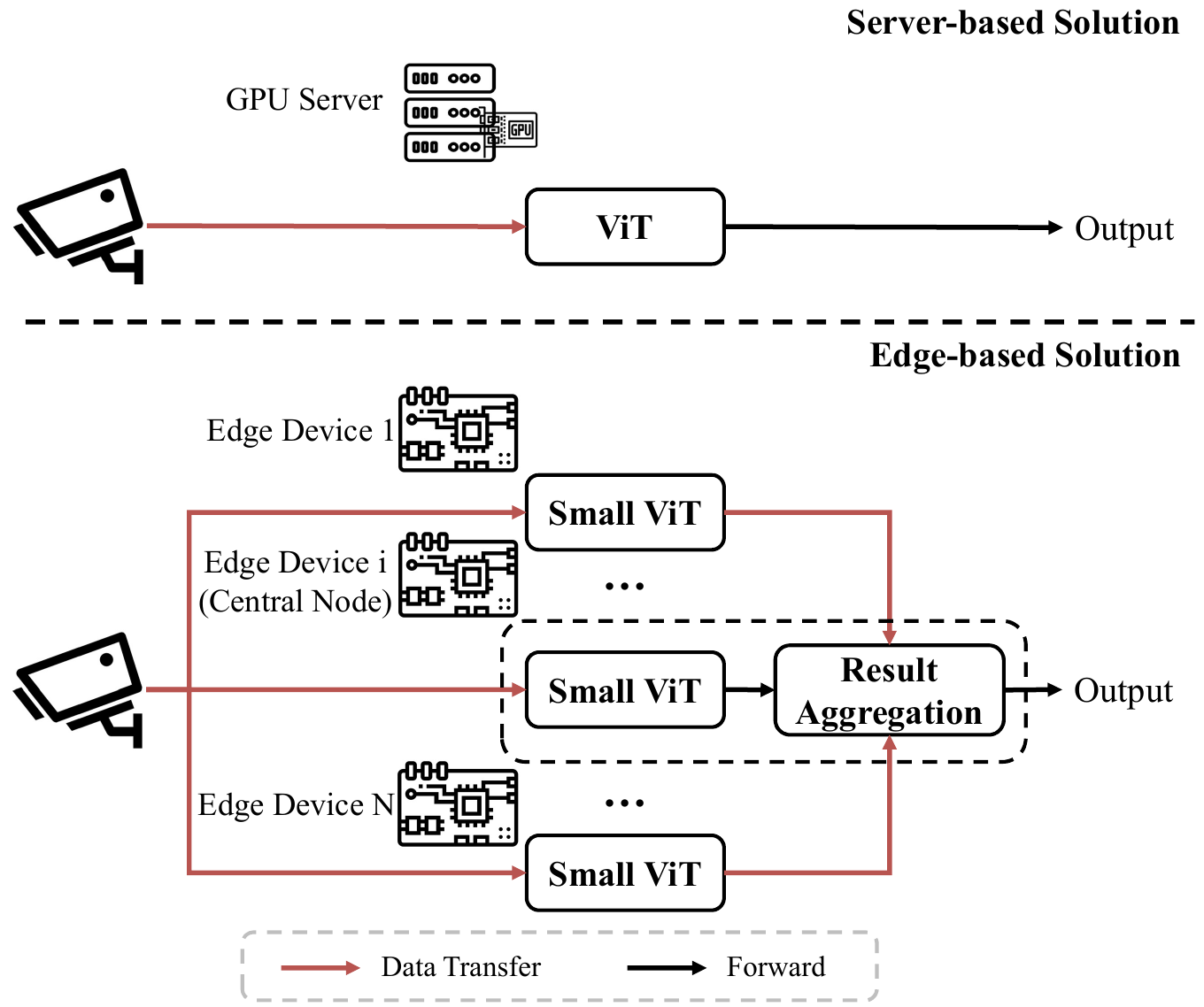}
    \caption{Comparison of server-based solution (top) and edge-based solution (bottom) for ViT deployment. 
    For sever-based solution, data need to be uploaded to the cloud. The GPU sever executes ViT inference. 
    For edge-based solution, it consists of multiple edge devices with deployed small ViT models. 
    Nearby edge devices execute inference after receiving data. 
    Then, the central device collects the intermediate results from all other devices and aggregates them to produce the inference output.}
    \label{fig:framework}
  \end{figure}
  
  \section{Collaborative Deployment of ViTs on Multiple Edge Devices}
  \label{sec:system}
  In this section, we design a collaborative deployment framework of ViT on multiple edge devices. 
  We first introduce the traditional deployment solution on the server. 
  Then we present the workflow of our framework and depict the details of each module in our framework.
  
  \subsection{System Structure and Workflow}
  In light of the high inference latency and energy consumption, 
  ViTs are usually deployed on commodity GPU servers, as shown in Fig. \ref{fig:framework} (top). 
  The workflow of such server-based solution is: 
  initially a giant \textit{ViT} model is deployed on a GPU server in a cloud data center; 
  then the input data (e.g., images, videos) from remote users are uploaded to the cloud; 
  the GPU server runs the ViT model for inference and transmits the results back to the users.
  This method is convenient to deploy but leads to high transmission cost and latency from the remote users to the cloud. 
  
  To tackle this issue, we propose the idea that data are directly transmitted to nearby edge devices to reduce the communication cost, 
  where these devices are collaboratively deployed for ViTs to improve inference performance. 
  This is achieved by the designed collaborative deployment framework of ViTs on multiple edge devices as shown in Fig. \ref{fig:framework} (bottom). 
  It contains multiple edge devices as well as a central node. 
  All the devices collaboratively conduct inference and can communicate with each other via a wireless network such as Wi-Fi, Zigbee, and Bluetooth. 
  The \textit{small ViTs} are deployed on edge devices. 
  When receiving an inference task, the collected data are first transmitted to all edge devices.
  These devices conduct inference with the \textit{small ViTs} in parallel. 
  Intermediate results of the \textit{small ViTs} from all other edge devices are transmitted to the central edge device, 
  where the \textit{result aggregation} module aggregates the intermediate results to produce the inference output.

  The aforementioned framework mainly depends on two modules: small ViT and result aggregation. 
  These modules are required to achieve the following goals:
  \begin{itemize}
    \item \textbf{Goal 1:} Small ViTs can be deployed on edge devices and achieve low inference latency. 
    \item \textbf{Goal 2:} The results of small ViTs can be aggregated to achieve a performance comparable to that of the powerful giant ViT.
  \end{itemize}

  \subsection{Decomposition and Aggregation}
  \label{subsec:decompose_aggregate}
  We illustrate how to decompose a giant ViT into small models and then aggregate intermediate results from small models to minimize accuracy loss.

  \textbf{Model Decomposition.} Considering the Goal 1, we propose to remove the redundant parts in the ViT to get small ViTs. 
  Clark \textit{et al.} \cite{bert_look} observed that particular heads correspond to specific semantic relations. 
  Michel \textit{et al.} \cite{16heads} found that performance are significantly affected by only a small quantity of heads. 
  Khakzar \textit{et al.} \cite{speific_neurons} also pointed out that plenty of neurons are unnecessary for specific tasks. 
  Hence, we can discard the redundant structures in the ViT according to the target task and hardware requirement, 
  and decompose the giant ViT into several smaller models. 
  Every decomposed small model is under the hardware and latency constraints. 
  The redundancy of ViT are mainly related to the embedding dimension, 
  the number of attention heads, the MLP dimension, and the number of network layers.
  We focus on these factors to conduct decomposition. 
  
\begin{figure}
    \centering
    \includegraphics[width=0.475\textwidth]{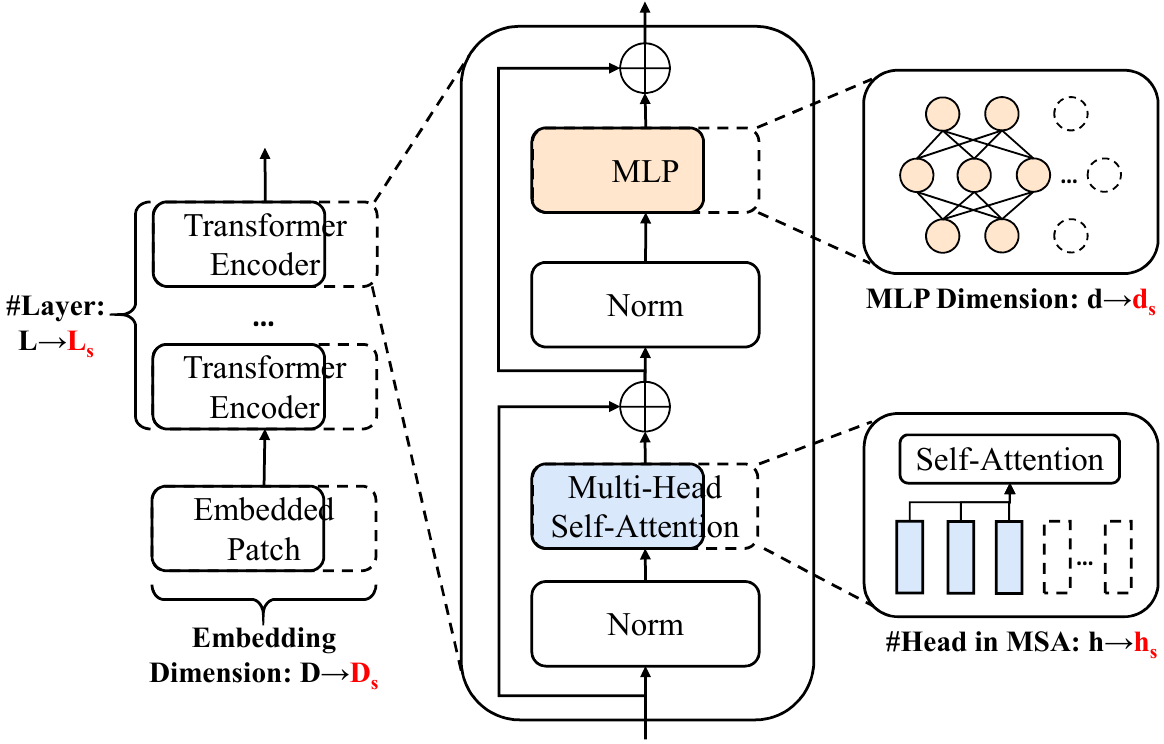}
    \caption{The process of model decomposition. 
    We focus on embedding dimension, the number of attention heads, the MLP dimension, 
    and the number of network layer when conducting decomposition. 
    The number of ViT layers is decreased from $L$ to $L_s$. 
    Then redundant heads in MSA and redundant neurons in MLP are removed. 
    In order to match feature dimension, the dimension of embeddings is also reduced to $D_s$. 
    The dimension of MLP is changed from $d$ to $d_s$.}
    \label{fig:decomposition}
  \end{figure}

  As shown in Fig. \ref{fig:decomposition}, the process of decomposition is as follows: 
  \begin{itemize}
    \item \textbf{Encoder reduction: }The number of transformer encoders $L$ is decreased to $L_s$. 
    \item \textbf{Head reduction: }Some redundant heads in the MSA module are removed and the reduced number of heads is $h_s$. 
    \item \textbf{Embedding reduction: }The embedding dimension is changed to $D_s=64h_s$, in order to match the feature dimension in the inference.
    \item \textbf{Neuron reduction: }Some redundant intermediate neurons in the MLP block are discarded and the dimension of MLP is changed from $d$ to $d_s$.
  \end{itemize}
  
  The reduction criterion will be discussed in Section \ref{subsec:shrinking}. 
  In order to avoid the case that some devices have completed inference but others have not, 
  we only focus on deployment of homogeneous devices in this work and make structures of all decomposed models the same. 

  \textbf{Feature Aggregation.} 
  Regarding the Goal 2, we propose to utilize ensemble methods to aggregate results. 
  Most ensemble methods\cite{DBLP:conf/cvpr/GuoWWYLHL20} only fuse predictions of different sub-models or select a reliable prediction. 
  However, the final results of these methods mainly depend on the performance of sub-models. 
  The ensemble model cannot achieve satisfactory performance due to the limited performance of sub-models. 
  Hence, we propose to break the performance restrictions of sub-models by aggregating the features of these models, 
  and design an efficient feature aggregation module $FA(\cdot)$ to fuse intermediate features. 
  In this module, the intermediate features from all edge devices are first concatenated, as: 
  \begin{equation}
    \label{equ:concate}
    \mathbf{X} = \text{Concat}\left(\mathbf{X}_1,\mathbf{X}_2,\ldots,\mathbf{X}_N\right),
  \end{equation}
  where $\mathbf{X}_i, i\in[1, N]$ are the intermediate features of small models. 
  The number of small models is $N$. 
  The concatenated feature $\mathbf{X} \in \mathbb{R}^{ND} $ is passed to an MLP block consisting of two fully-connected layers. 
  The linear transformation $\mathbf{W}_1 \in \mathbb{R}^{ND\times D_t}$ of the first fully-connected layer allows the interaction of different features.
  The transformed features can restore the initial dimension by utilizing the transformation $\mathbf{W}_2 \in \mathbb{R}^{D_t\times ND}$ of the second fully-connected layer.
  The feature aggregation module $FA(\cdot)$ can be formulated as:

  \begin{equation}
    \label{equ:fuse}
    \begin{split}
      \mathbf{X}_{fused} &= FA(\mathbf{X}_1,\mathbf{X}_2,\ldots,\mathbf{X}_N)\\
      &= \left(\mathbf{X}\mathbf{W}_1+\mathbf{b}_1\right)\mathbf{W}_2+ \mathbf{b}_2 .\\
    \end{split}
  \end{equation}

  We remove the original activation function in the MLP block. 
  This can help decrease the inference latency dramatically and incur almost no accuracy loss. 
  The results in Section \ref{sec:ablation} show that our aggregation module can guarantee a satisfactory performance 
  and only increase extremely low inference latency compared to other typical aggregation approaches.
  
  \textbf{Discussion.} 
  The proposed collaborative deployment framework of ViTs on multiple edge devices is specially designed to achieve efficient deployment of ViTs 
  so that the inference latency is low and accuracy is comparable to the powerful giant ViTs. 
  We propose to decompose the single giant transformer model to multiple small models, 
  and then deploy these small models on edge devices. 
  In this way, only small amounts of data need to be transmitted and feature aggregation module is efficient. 
  By decomposing ViT into models as small as possible and designing an efficient module to aggregate, 
  we can significantly reduce the inference latency and save plenty of energy with only a slight accuracy loss.

\section{Algorithm Design}
\label{sec:algorithm}

In order to decompose the ViT into lightweight models and aggregate results of small models to obtain a satisfactory performance, 
we propose decomposition and ensemble based on knowledge distillation termed DEKD.
In this section, we first provide an overview of our method, 
and then depict the details of each part.

\subsection{Overview}

\begin{figure*}
  \centering
  \includegraphics[width=0.92\textwidth]{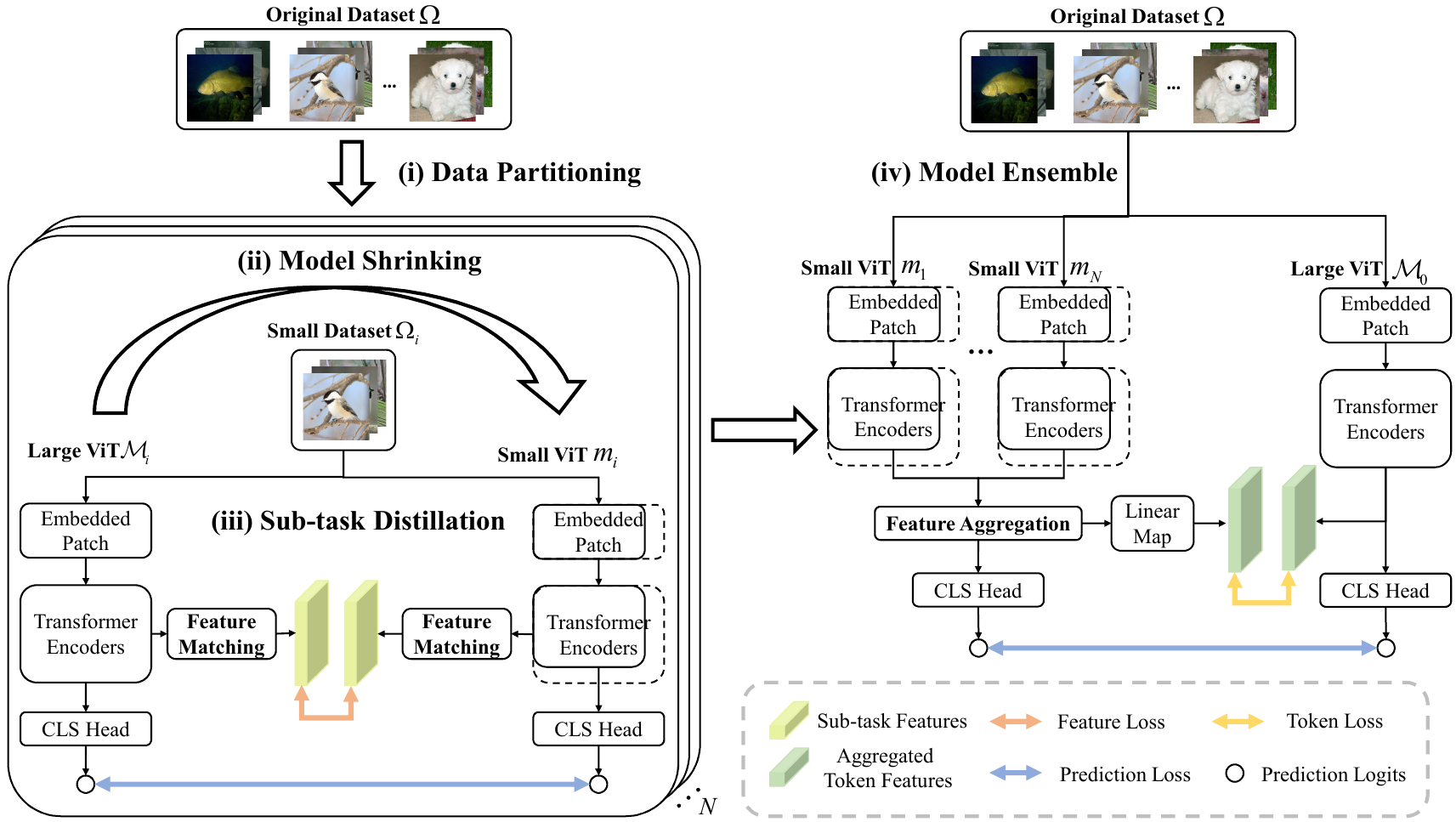}
  \caption{Overview of the DEKD algorithm. 
  The algorithm requires a dataset and multiple small datasets, 
  utilizes multiple large ViT as teacher models and multiple small ViT as student models, 
  as well as contains a feature matching module and aggregation module. 
  The procedure includes four steps: 
  (i) The original dataset is equally partitioned into $N$ small datasets. The large ViT is also decomposed into $N$ small models. 
  (ii) By model shrinking, unimportant parts for specific classes in the small models can be removed. 
  (iii) During sub-task distillation, the small models, as student, can learn fine-grained knowledge from the large teacher ViT by feature matching. 
  (iv) The $[class]$ tokens from the small models are fused by the feature aggregation module.  
  We also utilize the large ViT to improve the collaborative inference performance. 
  The aggregated results are passed into the task-specific head to obtain the final results.  }
  \label{fig:algorithm}
\end{figure*}

An overview of our proposed DEKD algorithm is shown in Fig. \ref{fig:algorithm}.
The algorithm requires a dataset and multiple small datasets, 
utilizes multiple large ViT as teacher models and multiple small ViT as student models, 
as well as contains a feature matching module and aggregation module. 
Assume we have $N$ edge devices, 
and the DEKD algorithm follows a four-step procedure as shown in Algorithm \ref{algo:dekd}:

(i) \textbf{Data partitioning}: 
We partition the original dataset into $N$ small datasets according label classes.  

(ii) \textbf{Model shrinking}: 
We also decompose the large ViT into $N$ small models.
Every small model only preserves the structure which is more important to the specific small dataset than other small datasets.

(iii) \textbf{Sub-task distillation}: 
We use $N$ small datasets to train $N$ large ViTs, which are randomly initialized, respectively. 
These pretrained large ViTs are regarded as teacher models. 
Each small model can learn the intermediate knowledge from the specific teacher model by utilizing the feature matching module.

(iv) \textbf{Model ensemble}: 
We utilize the feature aggregation module to fuse the $[class]$ tokens from the trained small models.  
Moreover, intermediate knowledge of the large ViT trained on the original dataset can also be used as supervision to improve the aggregation performance.

\subsection{Data Partitioning}
\label{subsec:partition}
We first introduce the proposed data partitioning approach as shown in Lines 1 to 4 of Algorithm \ref{algo:dekd}. 
Due to their limited model capacity, 
the small models cannot finish complicated inference tasks with an acceptable accuracy and latency. 
Thus, we propose to decompose the original dataset into multiple small datasets to simplify the tasks. 
Here we take an $\mathcal{N}_{class}$-class classification task as an example. 

First, the original dataset $\Omega_0  $ is equally decomposed into $N$ small datasets according to the class label. 
The number of class in the dataset is denoted by $\mathcal{N}_{class}$. 
If the number of class $\mathcal{N}_{class}$ is divisible by the number of small datasets $N$,
the number of class $\mathcal{N}_i$ in the small datasets is $\mathcal{N}_{class}/N$. 
Otherwise, the number of class $\mathcal{N}_i$ in the $i$-th small dataset is given by
\begin{equation}
  \label{equ:class_num}
  \mathcal{N}_i=\begin{cases}
    \lfloor \mathcal{N}_{class}/N \rfloor+1 \text{,} &\mbox{if \textit{i} = 1, 2, \ldots, \textit{n}}\\
    \lfloor \mathcal{N}_{class}/N \rfloor \text{,} &\mbox{if \textit{i = n+}1, \ldots, \textit{N}} 
  \end{cases}
\end{equation}
where $n=\mathcal{N}_{class} \text{ MOD } N$. 
Then, we randomly sample $\mathcal{N}_i$ classes of data from dataset $\Omega_0  $. 
These sampled data construct the small dataset $\Omega _i$. 
Finally, after repeating sampling and dataset construction for $N$ times, 
we obtain $N$ partitioned small datasets.

\begin{algorithm}[!t]
  \label{algo:dekd}
  \caption{DEKD}
  \DontPrintSemicolon

    \KwNotations{large ViT $\mathcal{M}_0$ trained on original dataset $\Omega_0  $; 
    large ViTs $\{\mathcal{M}_1,\ldots,\mathcal{M}_N\}$ trained on small datasets $\{\Omega_1,\ldots,\Omega_N\}$, respectively;
    shrinking factor $\sigma$;
    In sub-task distillation, learning rates $ \{\mu_1,\ldots,\mu_N\} $; 
    first momentum factors $\{\rho _{1,1},\ldots,\rho _{1,N}\}$;
    second momentum factors $\{\rho _{2,1},\ldots,\rho _{2,N}\}$
    weight decay factors $\{\lambda_1,\ldots,\lambda_N\}$;
    In model ensemble, 
    learning rate $ \mu $; 
    first momentum factor $\rho _1$;
    second momentum factor $\rho _2$; 
    weight decay factor $\lambda$; the ensemble of small trained models $m_0$;}

    \tcc{(1) Data Partition}

    Compute the number of class $\mathcal{N}_i$ by (\ref{equ:class_num});
      
    \For(){$i:=1$ to $N$}{
      Randomly sample $\mathcal{N}_i$ classes data from $\Omega_0  $;\\
      Using sampled data to generate small dataset $\Omega _i$;\\
      \tcc{(2) Model Shrinking}
      Using small dataset $\Omega _i$ to train $\mathcal{M}_i$;\\
      \tcc{(i) Rank by importance}
      Compute head importance $I_{h,i}$ for $\mathcal{M}_i$ by (\ref{equ:head_imp});\\
      Compute neuron importance $I_{w,i}$ for $\mathcal{M}_i$ by (\ref{equ:neuron_imp});\\
      Rank heads and neurons in descending order of importance;\\      
      \tcc{(ii) Remove unimportant parts}
      Remove last $\sigma$ percent of heads and neurons;\\
      Reconstruct the connections of the model and get new small model $m_i$;
      
    }

    \tcc{(3) Sub-task Distillation}
    \For(){$i:=1$ to $N$} 
    {
      $\bm{\theta}_i\leftarrow $ \textbf{Sub-task Distillation}$\left(\Omega _i, m_i, \mathcal{M}_i, \mu_i, \rho _{1,i}, \rho _{2,i}, \lambda_i\right)$
      
      } 
    \tcc{(4) Model Ensemble}

    \textbf{Initialize }parameters for $m_0$ using $\{\bm{\theta}_1,\ldots, \bm{\theta}_N\}$;\\
    \textbf{Initialize }parameters for large ViT $\mathcal{M}_0$ by training on original dataset $\Omega_0$;\\
    $\bm{\theta}_{0} \leftarrow $ \textbf{Model Ensemble}$\left(\Omega_0, \bm{\theta}_{0}, \mu, \rho_1, \rho_2, \lambda, \mathcal{M}_0\right)$;
    
\end{algorithm}

\subsection{Model Shrinking}
\label{subsec:shrinking}
When the small datasets are available, we next introduce the proposed model shrinking method. 
The ViT is decomposed into $N$ small models by following the process in Section \ref{subsec:decompose_aggregate}. 
In order to ensure only slight performance loss and simultaneously reduce the computation cost, 
we only preserve the most important parts in the small models. 
In the multiheaded self-attention module, different heads can capture various semantic information\cite{analyze_msa}. 
Some researchers\cite{speific_neurons} also found that some neurons are responsible for specific classes.
Thus, we can remove the unimportant heads in MSA and the unimportant neurons in the hidden layer of MLP to reduce the size and computation of the small models. 
The important parts for specific classes in the small models shall be reserved. 
We define the shrinking factor $\sigma$ to dynamically adjust the model size according to hardware requirements. 
The shrinking factor indicates that the last $\sigma \%$ of heads and neurons shall be removed. 

\begin{figure}[!h]
  \centering
  \includegraphics[width=0.48\textwidth]{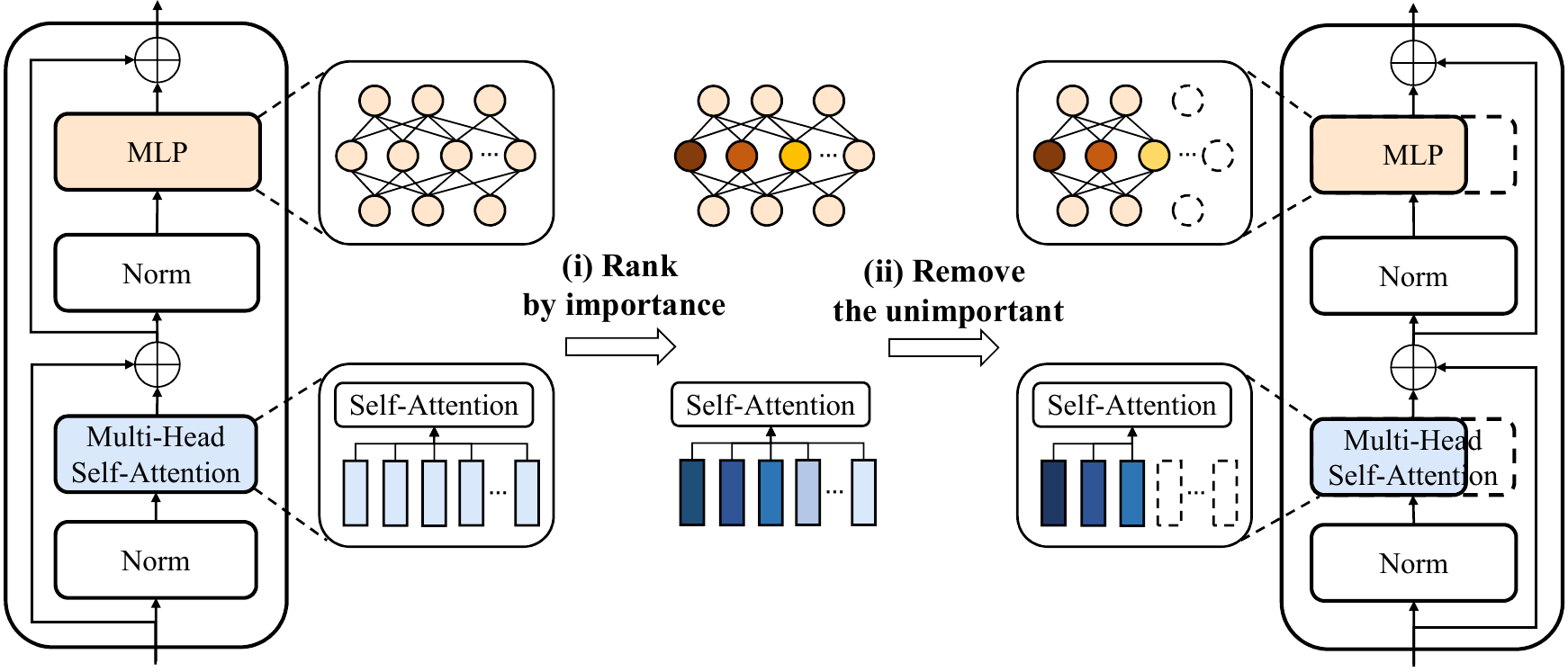}
  \caption{The procedure of Model Shrinking. 
  Firstly, the importance of each head in MSA and each neuron in the hidden layer of MLP is computed. 
  Then we sort heads and neurons by importance. 
  The unimportant parts for specific classes are then removed. 
  Finally, connections of the remaining heads and neurons are reconstructed.}
  \label{fig:criterion}
\end{figure}

Fig. \ref{fig:criterion} illustrates the details of the proposed model shrinking approach.
First, we use testing set of the small datasets partitioned from the original dataset to compute the importance of heads in the MSA module and of the neurons in the hidden layer of the MLP block.
Then, we rank these heads and neurons by their importance and remove unimportant parts. 
Finally, the connections of the small models are reconstructed. 

In order to identify the importance of heads and neurons in encoders for certain classes, 
we utilize the importance metric as in \cite{analyze_msa,DynaBERT}. 
The importance metric $I $ of a head or a neuron can be computed by using the variation of training loss $\mathcal{L}$ if a head or a neuron is removed.
For one head with output $\mathbf{O}_h$, the training loss $\mathcal{L}(\mathbf{O}_h)$ can be expressed as
\begin{equation}
  \mathcal{L}(\mathbf{O}_h)=\mathcal{L}(\mathbf{O}_h^{*})+\frac{\partial \mathcal{L}}{\partial \mathbf{O}_h}\left(\mathbf{O}_h-\mathbf{O}_h^{*}\right)+R_h,
\end{equation}
by using Taylor expansion, where $R_h$ is the remainder of the term that is smaller in magnitude than the previous terms if $\mathbf{O}_h$ is sufficiently close to $\mathbf{O}_h^{*}$.
If we remove head $h$, $\mathbf{O}_h^{*}=0$. Then, the importance of head $h$ is
\begin{equation}
  \label{equ:head_imp}
  \begin{split}
    I_{h}&=|\mathcal{L}(\mathbf{O}_h)-\mathcal{L}(\mathbf{O}_h=0)|\\
    &=\left| \frac{\partial \mathcal{L}}{\partial \mathbf{O}_h}\mathbf{O}_h+R_h \right|\approx \left| \frac{\partial \mathcal{L}}{\partial \mathbf{O}_h}\mathbf{O}_h \right|,
  \end{split}
\end{equation}
where the remainder $R_h$ can be ignored. 
Similarly, the weights of a neuron connected with $\mathbf{W}_1$ and $\mathbf{W}_2$ in the MLP block are defined as $w=\{w_1,w_2,\ldots,w_{2D}\}$.
Thus, the importance of the neuron is given by
\begin{equation}
  \label{equ:neuron_imp}
  I_{w}=\left| \frac{\partial \mathcal{L}}{\partial \mathbf{O}_w}\mathbf{O}_w \right|=\left| \sum_{i = 1}^{2D}   \frac{\partial \mathcal{L}}{\partial \mathbf{O}_{w_i}}\mathbf{O}_{w_i} \right|.
\end{equation}

\textbf{Discussion. }We use the small datasets partitioned from the original dataset to find the important parts for certain classes in the decomposed small models.
We can flexibly control the size by removing the unimportant parts in the MSA and MLP modules according to hardware requirements. 
Our method reduces the model size and redundant information at the same time to guarantees the trade-off between inference accuracy and latency. 
We summarized the procedure in Lines 5 to 10 of Algorithm \ref{algo:dekd}. 

In order to minimize accuracy loss caused by decomposition and shrinking, 
we propose the sub-task distillation approach to transfer knowledge for specific classes from large ViTs to small decomposed models.
The approach requires multiple small datasets, 
and contains multiple teachers and students as shown in the lower left part of Fig. \ref{fig:algorithm}. 
The teachers $\{\mathcal{M}_1, \mathcal{M}_2\ldots,\mathcal{M}_N\}$ are large ViTs trained on small datasets, respectively. 
The students are decomposed small models $m_i$. 
A small dataset has correspondence between a pair of teacher and student. 
The student can learn knowledge for specific classes with the help of the teacher trained on the corresponding small dataset. 
During distillation, the predictions and intermediate features in teachers are chosen to transfer. 
For the sake of intermediate distillation, 
we utilize a feature matching module to match the dimensions of teachers and students. 
In the following, we introduce the feature matching module, and then present the objective function for distilling the knowledge of large ViTs.

\textbf{Feature Matching. }
The proposed feature matching module $FM(\cdot)$ applies transformation in intermediate features to promote distillation when the feature dimensions of teachers and students are different. 
Recent works found that transferring knowledge base on features\cite{TinyBERT} can be utilized to help the training of the student. 
However, these methods usually have restrictions on mode layers\cite{mobilebert} or significant information loss when performing dimension matching \cite{minilmv2}. 
This motivates us to design a novel feature matching module to break the restriction and reduce the performance loss in distillation. 

\begin{figure}
  \centering
  \includegraphics[width=0.475\textwidth]{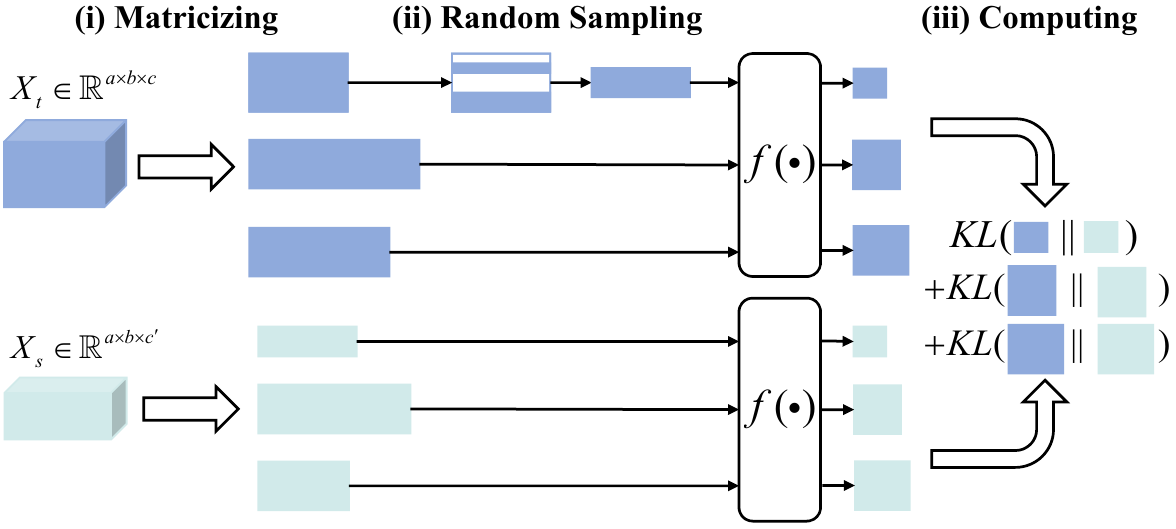}
  \caption{The feature matching module in sub-task distillation. 
  The 3D features from the teacher and student are matricized to be 2D matrices with different dimensions. 
  If dimensions of the matrices between the teacher and student are not the same, 
  the matrix from the teacher is randomly sampled to get the same number of rows as the matrix of the student. 
  Then these matrices are transferred to square matrices with the same dimension by $f(\cdot)$. 
  Finally, KL divergence can be used to calculate their differences.}
  \label{fig:match_feature}
\end{figure}

Details of our feature matching module are shown in Fig. \ref{fig:match_feature}. 
We suppose the feature map from teachers is $X_t \in \mathbb{R}^{a\times b\times c}$, 
and the feature map from students is $X_s \in \mathbb{R}^{a\times b\times c' }(c'\leq c)$. 
Firstly, the feature maps are expanded to be of three dimensional. 
By matricizing teacher and student according to each of their three dimensions, 
we can get six matrices $X_{t_1}\in \mathbb{R}^{a\times bc}, X_{t_2}\in \mathbb{R}^{b\times ac}, X_{t_3}\in \mathbb{R}^{c\times ab}, 
X_{s_1}\in \mathbb{R}^{a\times bc'}, X_{s_2}\in \mathbb{R}^{b\times ac'}$, and $X_{s_3}\in \mathbb{R}^{c'\times ab}$. 
Then, we utilize the function $f(X)=XX^{T}$ to match the second dimension of $X_{t_i}$ and $X_{s_i}$. 
But the numbers of the first dimension between $X_{t_3}$ and $X_{s_3}$ are different. 
Hence, before applying the function $f$, $X_{t_3}$ needs to be randomly sampled to get $c'$ rows to construct a new matrix $X'_{t_3}$. 
In this way, our feature matching module can break the restriction of feature dimension difference and reduce computation cost during feature matching.

\subsection{Sub-task Distillation}
\label{subsec:sub-distill}

\begin{algorithm}[!t]
  \label{algo:distill}
  \caption{Sub-task Distillation}
  \DontPrintSemicolon
    
    \KwInput{partitioned small dataset $\Omega _i$; 
    small model $m_i $; 
    large ViT models $\mathcal{M}_i$; 
    learning rates $ \mu_i $; 
    first momentum factors $\rho _{1,i} $; 
    second momentum factors $\rho _{2,i} $; 
    weight decay factors $\lambda_i $; 
    }
    \KwOutput{model $m_i$ with optimized parameters $\bm{\theta}_t$; }
      
    \textbf{Initialize } 
    the parameter $\bm{\theta}_{t=0}$ of model $m_i$  by training in pretraining dataset;\\
    \textbf{Initialize }
    time step $t\leftarrow 0$, 
    first moment vector $\bm{u}_{t=0}\leftarrow \bm{0}$, 
    $\epsilon\leftarrow 10^{-8}$, 
    second moment vector $\bm{v}_{t=0}\leftarrow \bm{0}$, 
    schedule multiplier $\eta_{t=0} \in \mathbb{R}^n$;
    max training steps $I_{max}\leftarrow 10^5$;\\
    \For(){$t:=1$ to $I_{max}$}{
      Select batch from $\Omega_i$;\\
      Apply $FM(\cdot)$ transformation in $\mathbf{Q},\mathbf{K}$, and $\mathbf{V}$ vectors;\\
      Compute $\mathcal{L}_{\text{feat}}$ with (\ref{equ:feat_loss});\\
      Compute $\mathcal{L}_{\text{sub-task}}$ with (\ref{equ:sub_task_loss});\\
      $\bm{g}_t\leftarrow \mathcal{L}_{\text{sub-task}}(\bm{\theta}_{t-1})+\lambda_i\bm{\theta}_{t-1}$;\\
      $\bm{u}_t\leftarrow \rho _{1,i}\bm{u}_{t-1}+(1-\rho _{1,i})\bm{g}_t$;\\
      $\bm{v}_t\leftarrow \rho _{2,i}\bm{v}_{t-1}+(1-\rho _{2,i})\bm{g}_t^2 $;\\
      $\hat{\bm{u}}_t\leftarrow\bm{u}_t/(1-\rho _{1,i}^t)$;\\
      $\hat{\bm{v}}_t\leftarrow\bm{v}_t/(1-\rho _{2,i}^t) $;\\
      $\eta_t\leftarrow SetScheduleMultiplier(t)$;\\
      $\bm{\theta}_t\leftarrow\bm{\theta}_{t-1}-\eta_t\left(\mu_{i}\hat{\bm{u}}_t/(\sqrt[]{\hat{\bm{v}}_t}+\epsilon)+\lambda\bm{\theta}_{t-1}\right)$;\\
      \If{stopping criterion is met}
      {
        \textbf{Output } $m_i$ with optimized parameters $\bm{\theta}_t$;\\
        \textbf{break};\\
      }
    }
      
\end{algorithm}

\textbf{Feature and Prediction Distillation. }
We use the predictions and intermediate features of teachers to guide the training of students. 
The student can learn knowledge from the teacher by minimizing the following objective:
\begin{equation}
  \label{equ:sub_task_loss}
  \mathcal{L}_{\text{sub-task}}=\mathcal{L}_{pred}+\beta\mathcal{L}_{feat},
\end{equation}
where $\beta$ is a balancing hyperparameter. 

For distillation based on predictions, we follow \cite{deit} to take the hard decision of the teacher as a true label.
The prediction logits of teacher and student are $Y_t$ and $Y_s$, respectively.
Different from the traditional hard label in knowledge distillation, 
we utilize the decision of teacher as the label: 
\begin{equation}
  y_t=\argmax_{s} Y_t(s),
\end{equation} 
where $s$ represents all categories in the training set, 
and $y_t$ plays the same role as true label $y$, but with a better performance than $y$. 
The hard decision of teacher may change depending on the specific data augmentation. 
The student can learn more knowledge with data augmentation especially for the ViTs which require extensive data augmentations to guarantee performance. 
Hence, using $y_t$ is a better choice. 
The objective of prediction distillation is
\begin{equation}
  \label{equ:pred_loss}
  \mathcal{L}_{pred} = \frac{1}{2}\mathcal{L}_{CE}\left(\phi(Y_s),y \right)+\frac{1}{2}\mathcal{L}_{CE}\left(\phi(Y_s),y_t \right),
\end{equation}
where $\phi(\cdot)$ represents the prediction results and $\mathcal{L}_{CE}$ is the cross entropy loss, given by 
\begin{equation}
  \mathcal{L}_{CE}(p,q)=-\sum_{i = 1}^{} p_i\text{log}(q_i) . 
\end{equation}

In the distillation based on intermediate features, 
we utilize the $\mathbf{Q},\mathbf{K}$, and $\mathbf{V}$ vectors in MSA modules to supervise the training of students.
Recently, it is found that attention mechanism can be utilized to capture rich semantic information\cite{bert_look}.
Since the $\mathbf{Q},\mathbf{K}$, and $\mathbf{V}$ vectors are the key parts of MSA module, 
we utilize these vectors as the supervision of students to learn the semantic information of teachers. 
The dimensions of the three vectors are $h$, $M$ and $64$, respectively, 
where $h$ is the number of heads and $M$ is the number of tokens.
The number of heads and tokens are different between teachers and students due to the model decomposition and shrinking. 
Hence, the feature matching module $FM(\cdot)$ is utilized to match their dimensions. 
For notational simplicity, we denote $\mathbf{Q},\mathbf{K}$, and $\mathbf{V}$ as $\mathbf{P}_1, \mathbf{P}_2$, and $\mathbf{P}_3$, respectively.
The distillation objective is represented as
\begin{equation}
  \label{equ:feat_loss}
  \mathcal{L}_{feat}=\frac{1}{3l}\sum_{j,k = 1}^{l} \sum_{i = 1}^{3} \alpha_i \mathcal{L}_{KL}\left(FM(\mathbf{P}_{i,s}^{j}) || FM(\mathbf{P}_{i,t}^{k})\right), 
\end{equation}
\begin{equation}
  \mathcal{L}_{KL}(p||q)=\sum_{i = 1}^{} q_i\text{log}\frac{q_i}{p_i},
\end{equation}
where $\alpha_i$ is a balancing hyperparameter and $\mathcal{L}_{KL}$ is the KL-divergence loss to help learn the distribution of target samples.
Superscripts $j$ and $k$ signify the $j$-th layer of the student and the $k$-th layer of the teacher, respectively. 
The student model has $j$ layers, 
and we use a linear mapping function $g(\cdot)$ to select the layer of teachers to transfer, i.e., $k=g(j)$. 
For example, the $j$-th layer of the student model can learn knowledge from the $g(j)$-th layer of the teacher. 
By utilizing the mapping function for layers, we can transfer knowledge from teachers to the students layer-to-layer.

\textbf{Discussion. }
The proposed sub-task distillation method can help to transfer knowledge for specific classes from large ViT to decomposed small models. 
The predictions and intermediate features from teachers can promote the small models to learn more fine-grained knowledge. 
This can substantially reduce the performance drop caused by decomposition and shrinking. 
The sub-task distillation procedure is summarized in Algorithm \ref{algo:distill}. 

\begin{algorithm}[!ht]
  \label{algo:ensemble}
  \caption{Model Ensemble}
  \DontPrintSemicolon
    
    \KwInput{dataset $\Omega_0$; 
    the ensemble of small trained models $m_0$ with its parameters $\bm{\theta}_{0}$; 
    learning rate $ \mu $; 
    first momentum factor $\rho _1$;
    second momentum factor $\rho _2$; 
    weight decay factor $\lambda$;
    teacher model $\mathcal{M}_0$;}
    \KwOutput{model $m_0$ with optimized parameters $\bm{\theta}_t$;}
    
      \textbf{Initialize }
      model $m_0$ parameter $\bm{\theta}_{t=0} \leftarrow \bm{\theta}_{0}$, 
      time step $t\leftarrow 0$, $\epsilon\leftarrow 10^{-8}$, 
      first moment vector $\bm{u}_{t=0}\leftarrow \bm{0}$, 
      second moment vector $\bm{v}_{t=0}\leftarrow \bm{0}$, 
      schedule multiplier $\eta_{t=0} \in \mathbb{R}^n$;
      max training steps $I_{max}\leftarrow 10^5$;\\
      \For(){$t:=1$ to $I_{max}$}{
      Select batch from $\Omega_0$;\\
      Utilize $FA(\cdot)$ to aggregate CLS tokens with (\ref{equ:fuse});\\
      Compute $\mathcal{L}_{\text{token}}$ with (\ref{equ:token_loss});\\
      Compute $\mathcal{L}_{\text{ensemble}}$ with (\ref{equ:ensemble_loss});\\
      $\bm{g}_t\leftarrow \mathcal{L}_{\text{sub-task}}(\bm{\theta}_{t-1})+\lambda_i\bm{\theta}_{t-1}$;\\
      $\bm{u}_t\leftarrow \rho _{1,i}\bm{u}_{t-1}+(1-\rho _{1,i})\bm{g}_t$;\\
      $\bm{v}_t\leftarrow \rho _{2,i}\bm{v}_{t-1}+(1-\rho _{2,i})\bm{g}_t^2 $;\\
      $\hat{\bm{u}}_t\leftarrow\bm{u}_t/(1-\rho _{1,i}^t)$;\\
      $\hat{\bm{v}}_t\leftarrow\bm{v}_t/(1-\rho _{2,i}^t) $;\\
      $\eta_t\leftarrow SetScheduleMultiplier(t)$;\\
      $\bm{\theta}_t\leftarrow\bm{\theta}_{t-1}-\eta_t\left(\mu_{i}\hat{\bm{u}}_t/(\sqrt[]{\hat{\bm{v}}_t}+\epsilon)+\lambda\bm{\theta}_{t-1}\right)$;\\
      \If{stopping criterion is met}
        { 
          \textbf{Output } $m_0$ with optimized parameters $\bm{\theta}_t$;\\
          \textbf{break};\\
        }
    }
\end{algorithm}

\subsection{Model Ensemble}

After the distillation of small models using corresponding small datasets, 
we resort to a distillation-based model ensemble method to improve the collaborative inference of multiple small models. 
In this subsection, we depict the method and then formulate the objective function for training the multiple small models. 

The architecture of the method is shown in the right part of Fig. \ref{fig:algorithm}. 
The method is composed of a teacher and a student, and uses a dataset $\Omega_0 $. 
The teacher $\mathcal{M}_0$ is the large ViT trained on the original dataset $\Omega_0 $. 
The student is the ensemble of multiple decomposed small models trained by sub-task distillation. 
Based on knowledge distillation, 
the large ViT can contribute to the ensemble of multiple small models to achieve a good collaborative inference performance.

\textbf{Token and Prediction Distillation. }
We use the $[class]$ token and predictions from the teacher for supervision of the student.
The objective function is given by
\begin{equation}
  \label{equ:ensemble_loss}
  \mathcal{L}_{\text{ensemble}}=\mathcal{L}_{pred}+\gamma \mathcal{L}_{token},
\end{equation}
where $\gamma$ is a balancing hyperparameter. 
$\mathcal{L}_{pred}$ and $\mathcal{L}_{token}$ are the objective functions for prediction-based and token-based distillation, respectively. 
Different from Section \ref{subsec:sub-distill} where student is the single small model, 
the student here is composed of multiple small models which are responsible for specific classes, respectively. 
Each small model can extract different features, 
and thus it is not appropriate to utilize the same intermediate features to supervise the student training as these in Section \ref{subsec:sub-distill}.
In the ViT model, the $[class]$ token, including target semantic information, is used to predict the class. 
Hence, considering the trade-off between ensemble performance and complexity, we aggregate the $[class]$ tokens of all small models. 
The objective function of token distillation is
\begin{equation}
  \label{equ:token_loss}
  \mathcal{L}_{token} = \mathcal{L}_{MSE}\left( FA(Z_s)G, Z_t \right),
\end{equation}
where $Z_s=\{Z_{s_1},Z_{s_2},\ldots,Z_{s_N} \in \mathbb{R} ^{1\times D_s}\}$ is the set of $[class]$ tokens of all small models, 
$Z_t\in \mathbb{R} ^{1\times D_t}$ is the $[class]$ token of the teacher, 
and the matrix $G \in \mathbb{R} ^{ND_s\times D_t}$ is a learnable linear map, 
which transforms the aggregated $[class]$ tokens of all small models into the space of the teacher. 
The feature aggregation module $FA(\cdot)$ has the same structure as the aggregation part in Section \ref{subsec:decompose_aggregate}.
The mean squared loss $\mathcal{L}_{MSE}$ is utilized to represent the differences between the aggregated tokens and the token from the teacher.
The objective function of prediction distillation is the same as the one adopted in (\ref{equ:pred_loss}).

\textbf{Discussion. }
We propose a distillation-based model ensemble method to fuse multiple small models and improve the collaborative inference performance. 
By using both token and prediction distillation, 
we guarantee slight accuracy loss compared to large ViT. 
The procedure of Model Ensemble is illustrated in Algorithm \ref{algo:ensemble}.

\begin{table}
  \centering
  \renewcommand{\arraystretch}{0.8}
  \setlength{\tabcolsep}{1.8mm}
  \setlength{\extrarowheight}{0pt}
  \addtolength{\extrarowheight}{\aboverulesep}
  \addtolength{\extrarowheight}{\belowrulesep}
  \setlength{\aboverulesep}{0pt}
  \setlength{\belowrulesep}{0pt}
  \caption{Configuration Settings of Different Backbones Utilized in Our Method}
  \label{config}
  \begin{tabular}{c|c|cccc} 
  \toprule
  \#                                  & \textbf{Model}     & \textbf{\#Layers} & \begin{tabular}[c]{@{}c@{}}\textbf{Embedding }\\\textbf{dimension}\end{tabular} & \textbf{MLP size} & \textbf{\#Heads}  \\ 
  \hline
  1                                   & ViT-L/16           & 24                & 1024                                                                            & 4096              & 16                \\
  \rowcolor[rgb]{0.949,0.949,0.949} 2 & \textbf{DeViT}     & 12                & 192                                                                             & 768               & 3                 \\ 
  \hline
  3                                   & DeiT-B             & 12                & 768                                                                             & 3072              & 12                \\
  \rowcolor[rgb]{0.949,0.949,0.949} 4 & \textbf{DeDeiT}    & 12                & 192                                                                             & 768               & 3                 \\ 
  \hline
  5                                   & CCT-14\_7$\times$2 & 14                & 384                                                                             & 1152              & 6                 \\
  \rowcolor[rgb]{0.949,0.949,0.949} 6 & \textbf{DeCCT}     & 7                 & 256                                                                             & 512               & 4                 \\
  \bottomrule
  \end{tabular}
  \vspace{-5pt}
  \end{table}

\section{Experiments}
\label{sec:experiment}
In this section, we first introduce some implementation details and experiment settings. Then we evaluate our methods on GPU servers and edge devices, respectively. 
Finally, we conduct extensive ablation experiments to demonstrate the superiority of our methods.

\begin{figure}[!t]
  \centering
  \includegraphics[width=0.38\textwidth]{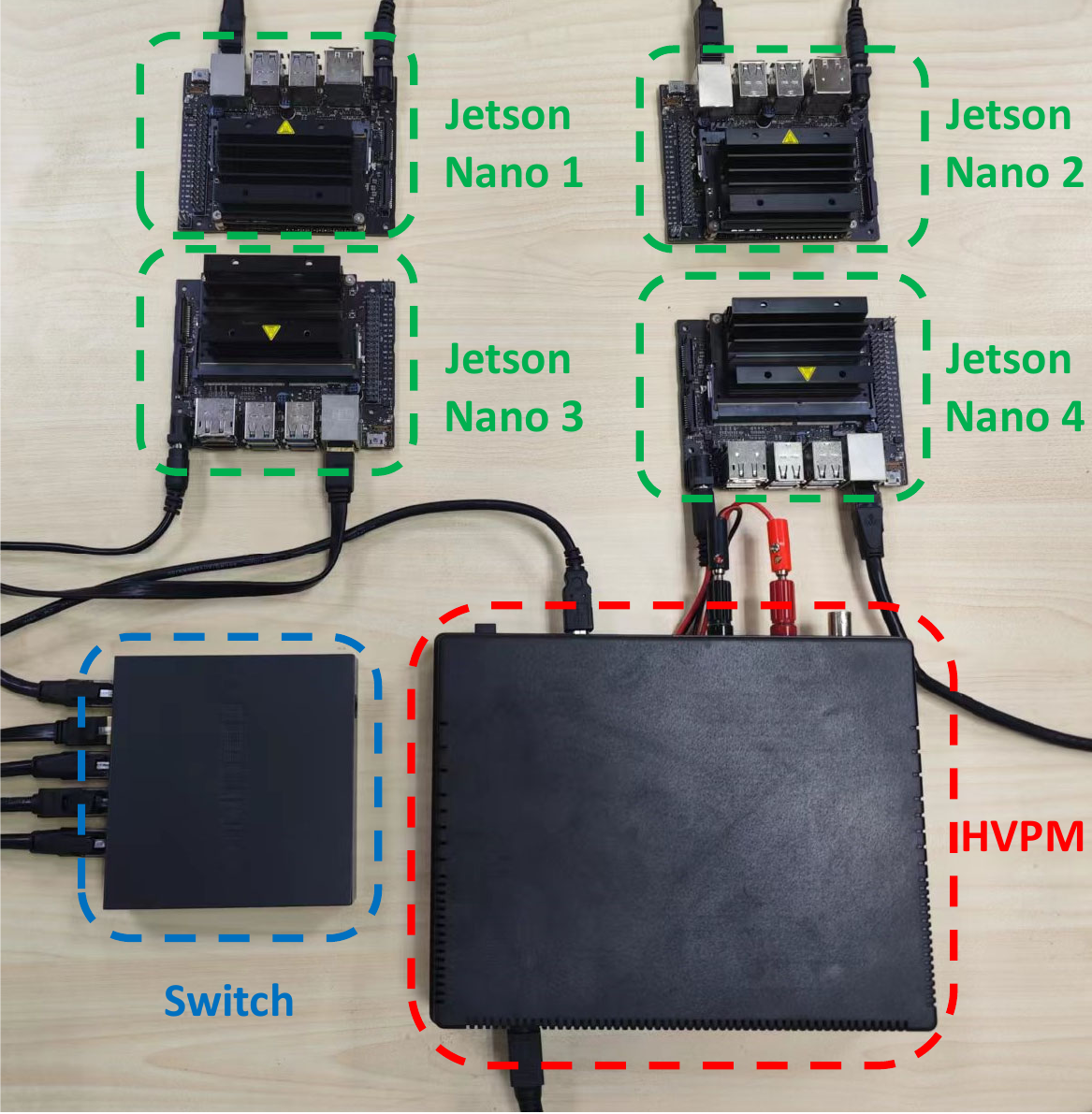}
  \caption{Our experimental prototype employs four Jetson Nano and one switch. 
  We utilize the Monsoon High Voltage Power Monitor (HVPM) to measure the energy and power.}
  \label{fig:devices}
  \vspace{-5pt}
\end{figure}

\begin{table*}
  \centering
  \renewcommand{\arraystretch}{0.9}
  \setlength{\tabcolsep}{1.7mm}
  \setlength{\extrarowheight}{0pt}
  \addtolength{\extrarowheight}{\aboverulesep}
  \addtolength{\extrarowheight}{\belowrulesep}
  \setlength{\aboverulesep}{0pt}
  \setlength{\belowrulesep}{0pt}
  \caption{Comparison Between Our Methods and Large ViTs in the GPU Server Using Different Datasets}
  \vspace{-5pt}
  \label{tab:result_gpu_large}
  \begin{tabular}{c|cc|cc|cc|cc|cc} 
  \toprule
  \multirow{2}{*}{\textbf{Model}}                    & \multirow{2}{*}{\textbf{\#Params.}} & \multirow{2}{*}{\textbf{FLOPs}} & \multicolumn{2}{c|}{\textbf{Results: CIFAR-100}}                               & \multicolumn{2}{c|}{\textbf{Results: Stanford Cars}}     & \multicolumn{2}{c|}{\textbf{Results: 102 Flowers}}       & \multicolumn{2}{c}{\textbf{Results: ImageNet-1K}}         \\
                                                      &                                      &                                  & \textbf{Acc.~$\uparrow$} & \textbf{Latency~$\downarrow$}                       & \textbf{Acc.~}$\uparrow$ & \textbf{Latency~$\downarrow$~} & \textbf{Acc.~$\uparrow$} & \textbf{Latency~$\downarrow$} & \textbf{Acc.~}$\uparrow$ & \textbf{Latency~$\downarrow$}  \\ 
  \hline
  ViT-L/16                                           & 304.0 M                                & 190.7 G                          & 88.77 \%                  & 18.65{\tiny$\pm$0.5} ms                                    & 89.83 \%                 & 18.68{\tiny$\pm$0.8} ms               & 98.13 \%                 & 18.19{\tiny$\pm$1.5} ms              & 76.50 \%                 & 18.57{\tiny$\pm$0.7} ms               \\
  \rowcolor[rgb]{0.949,0.949,0.949} \textbf{DeViTs}  & 23.0~M                              & 5.9~G                           & 87.98 \%                  & \textbf{6.45}{\tiny$\pm$0.3} ms                   & 89.11 \%                 & \textbf{6.25}{\tiny$\pm$0.7} ms      & 96.73 \%                 & \textbf{6.50}{\tiny$\pm$0.2} ms                & 74.85 \%                 & \textbf{6.16}{\tiny$\pm$0.3} ms       \\ 
  \hline
  DeiT-B                                             & 86.6~M                              & 35.1~G                          & 90.80 \%                  & 8.51{\tiny$\pm$0.2} ms                                     & 92.10 \%                 & 8.62{\tiny$\pm$0.8} ms                & 98.40 \%                 & 8.39{\tiny$\pm$0.1} ms               & 83.40 \%                 & 8.32{\tiny$\pm$0.7} ms                \\
  \rowcolor[rgb]{0.949,0.949,0.949} \textbf{DeDeiTs} & 23.9~M                              & 5.9~G                           & 89.34 \%                  & \textbf{6.62}{\tiny$\pm$0.3} ms                   & 91.53 \%                 & \textbf{6.51}{\tiny$\pm$0.4} ms       & 97.89 \%                 & \textbf{6.76}{\tiny$\pm$0.6} ms               & 81.94 \%                 & \textbf{6.44}{\tiny$\pm$0.3} ms       \\ 
  \hline
  CCT-14\_7$\times$2                                 & 22.4~M                              & 11.1~G                          & 88.78 \%                  & 8.56{\tiny$\pm$0.2} ms                                     & 91.78 \%                 & 9.11{\tiny$\pm$0.3} ms                & 97.97 \%                 & 10.16{\tiny$\pm$1.1} ms              & 80.67 \%                 & 8.96{\tiny$\pm$0.1} ms                \\
  \rowcolor[rgb]{0.949,0.949,0.949} \textbf{DeCCTs}  & 18.3~M                              & 5.1~G                           & 87.43 \%                  & \textbf{5.16}{\tiny$\pm$0.2} ms & 90.67 \%                 & \textbf{5.35}{\tiny$\pm$0.3} ms       & 96.84 \%                 & \textbf{5.30}{\tiny$\pm$0.1} ms                & 79.12 \%                 & \textbf{5.38}{\tiny$\pm$0.2} ms       \\
  \bottomrule
  \end{tabular}
  \end{table*}

\begin{table*}
  \centering
  \renewcommand{\arraystretch}{0.9}
  \setlength{\tabcolsep}{1.6mm}
  \setlength{\extrarowheight}{0pt}
  \addtolength{\extrarowheight}{\aboverulesep}
  \addtolength{\extrarowheight}{\belowrulesep}
  \setlength{\aboverulesep}{0pt}
  \setlength{\belowrulesep}{0pt}
  \caption{Comparison Between Our Methods and Lightweight Baseline Models in the GPU Server Using Different Datasets}
  \vspace{-5pt}
  \label{tab:result_gpu_small}
  \begin{tabular}{c|cc|cc|cc|cc|cc} 
  \toprule
  \multirow{2}{*}{\textbf{Model}}                    & \multirow{2}{*}{\textbf{\#Params.}} & \multirow{2}{*}{\textbf{FLOPs}} & \multicolumn{2}{c|}{\textbf{Results: CIFAR-100}}         & \multicolumn{2}{c|}{\textbf{Results: Stanford Cars}}     & \multicolumn{2}{c|}{\textbf{Results: 102 Flower}}        & \multicolumn{2}{c}{\textbf{Results: ImageNet-1K}}         \\
                                                      &                                     &                                 & \textbf{Acc.} $\uparrow$ & \textbf{Latency} $\downarrow$ & \textbf{Acc.}~$\uparrow$ & \textbf{Latency} $\downarrow$ & \textbf{Acc. }$\uparrow$ & \textbf{Latency} $\downarrow$ & \textbf{Acc.} $\uparrow$ & \textbf{Latency} $\downarrow$  \\ 
  \hline
  T2T-ViT$_t$-14                                     & 21.5 M                              & 12.2 G                          & 86.62 \%                 & 9.89{\tiny$\pm$0.8} ms               & 90.98~\%                 & 10.38{\tiny$\pm$0.9} ms              & 97.81 \%                 & 9.64{\tiny$\pm$0.1} ms               & 81.70 \%                 & 9.84{\tiny$\pm$0.7} ms                \\
  Twins-SVT-S                                        & 24.0 M                              & 2.9 G                           & 86.72~\%                 & 13.79{\tiny$\pm$0.9} ms              & 91.08~\%                 & 14.84{\tiny$\pm$1.3} ms              & 97.82 \%                 & 15.78{\tiny$\pm$1.5} ms              & 81.70 \%                 & 15.18{\tiny$\pm$1.4} ms               \\
  Twins-PCPVT-S                                      & 24.1 M                              & 3.8 G                           & 86.70~\%                 & 13.13{\tiny$\pm$1.3} ms              & 91.46~\%                 & 14.33{\tiny$\pm$1.3} ms              & \textbf{97.98} \%        & 13.01{\tiny$\pm$0.1} ms              & 81.20 \%                 & 13.85{\tiny$\pm$1.0} ms               \\
  MobileViT-S                                        & 5.0 M                               & 2.0 G                           & 87.00~\%                 & 9.22{\tiny$\pm$0.7} ms               & 89.99~\%                 & 9.49{\tiny$\pm$0.2} ms               & 97.63 \%                 & 9.29{\tiny$\pm$0.1} ms               & 78.40 \%                 & 9.40{\tiny$\pm$0.1} ms                \\ 
  MobileViT-S-Ens                                    & 20.1 M                              & 8.1 G                           & 85.80~\%                 & 9.72{\tiny$\pm$0.6} ms               & 88.12~\%                 & 9.81{\tiny$\pm$0.3} ms               & 96.01 \%                 & 9.96{\tiny$\pm$0.2} ms               & 78.62 \%                 & 9.98{\tiny$\pm$0.2} ms                \\ 
  \hline\hline
  \rowcolor[rgb]{0.949,0.949,0.949} \textbf{DeViTs}  & 23.0 M                              & 5.9 G                           & 87.98~\%                 & 6.45{\tiny$\pm$0.3} ms               & 89.11~\%                 & 6.25{\tiny$\pm$0.7} ms               & 96.76 \%                 & 6.50{\tiny$\pm$0.2} ms               & 74.85 \%                 & 6.16{\tiny$\pm$0.3} ms                \\
  \rowcolor[rgb]{0.949,0.949,0.949} \textbf{DeDeiTs} & 23.9 M                              & 5.9 G                           & \textbf{89.34~}\%        & 6.62{\tiny$\pm$0.3}~ms               & \textbf{91.53~}\%        & 6.51{\tiny$\pm$0.4} ms               & 97.89 \%                 & 6.76{\tiny$\pm$0.6} ms               & \textbf{81.94} \%        & 6.44{\tiny$\pm$0.3} ms                \\
  \rowcolor[rgb]{0.949,0.949,0.949} \textbf{DeCCTs}  & 18.3 M                              & 5.1 G                           & 87.43~\%                 & \textbf{5.16}{\tiny$\pm$0.2}~ms      & 90.67~\%                 & \textbf{5.35}{\tiny$\pm$0.3} ms      & 96.84 \%                 & \textbf{5.30}{\tiny$\pm$0.1} ms      & 79.12 \%                 & \textbf{5.38}{\tiny$\pm$0.2} ms       \\
  \bottomrule
  \end{tabular}
  \end{table*}

\subsection{Implementation Details}
In this subsection, we first introduce prototype and deployment details. 
Then we depict the datasets, backbones, and counterparts, followed by evaluation metrics.

\textbf{Prototype.} 
We implement DeViT on a real-world collaborative edge computing testbed that consists of 4 NVIDIA Jetson Nano devices \cite{jetson_nano} and 1 switch. 
Figure \ref{fig:devices} shows the implemented hardware platform for DeViT. 
The Jetson Nano contains a Quad-core ARM Cortex-A57 MPCore processor at 1.6 GHz, a 4 GB RAM Memory, which represents the weak edge device. 
These edge devices are connected via a gigabyte switch TP-LINK TL-SG1008D. 
For bandwidth control, we use the traffic control tool tc \cite{tc}, which is able to limit the bandwidth under the setting value. 
The maximum bandwidth between devices are fixed at 2 MB/s.

\textbf{Deployment.}
We use PyTorch as the backend engine to execute ViT models. The employed ViT models are trained following DEKD algorithm and deployed on all devices in advance. 
The communication module is implemented based on gPRC \cite{grpc}. All the compared approaches are run with timm \cite{timm}, a GitHub repository, for fair comparison.

\textbf{Datasets. }
We evaluate the proposed methods on four public computer vision datasets: CIFAR-100 \cite{cifar}, 
Oxford 102 Flower \cite{flowers102}, Stanford Cars \cite{stanford_cars}, and ImageNet-1K \cite{imagenet_1k}. 
All images are resized to be 224$\times$224 in our experiment. 
The details of each dataset are given as follows: 
\begin{itemize}
  \item \textbf{CIFAR-100} consists of 60,000 images that belong to 100 classes uniformly. 
  In each class, there are 500 images for training and 100 images for testing.
  \item \textbf{Oxford 102 Flower} is an image classification dataset consisting of 102 flower categories. 
  The flowers are common in the United Kingdom. Each class consists of 40 to 258 images.
  \item \textbf{Stanford Cars} dataset consists of 196 classes of cars with a total of 16,185 images, taken from the rear angle. 
  The dataset is divided into almost equal train/test subsets with 8,144 training images and 8,041 testing images. 
  \item 
  \textbf{ImageNet-1K} is a large-scale hand-labeled image recognition dataset consisting of 1,000 categories. 
  There are 1.2 million samples in the training set and 50,000 samples in the testing set.
\end{itemize}

\textbf{Backbones. }
The configurations of different backbones are listed in Table \ref{config}. 
We show the large models and their corresponding small models decomposed by our method.
In the table, \#1, \#3, and \#5 represent the large ViTs. 
\#2, \#4, and \#6 show the decomposed small models using our method. 
We use the following models as backbones: 
\begin{itemize}
  \item \textbf{ViT} \cite{vit}: The first pure transformer structure for computer vision tasks.
  \item \textbf{DeiT} \cite{deit}: DeiT adds a distillation token to learn knowledge of the teacher, as compared to ViT.
  \item \textbf{CCT} \cite{cct}: CCT is a CNN+Transformer hybrid model that uses CNNs and sequence pooling to replace patch embeddings and class token. 
\end{itemize}

\textbf{Counterparts. }
We first compare the large ViTs for aforementioned three backbones, including ViT-L/16, DeiT-B and CCT-14\_7$\times$2, to demonstrate the effectiveness of our methods. 
Then we compare our methods with four representative lightweight ViT approaches to show the superiority. 
\begin{itemize}
  \item \textbf{T2T-ViT$_t$-14 }\cite{t2tvit} incorporates a layer-wise transformation to 
  structuralize the image to tokens by recursively aggregating neighboring tokens. 
  \item \textbf{Twins-PCPVT-S }\cite{twins} introduces a carefully-designed yet simple spatial attention mechanism. 
  \item \textbf{Twins-SVT-S }\cite{twins} introduces locally-grouped self-attention and global sub-sampled attention to capture fine-grained and global information. 
  \item \textbf{MobileViT-S }\cite{mobile_vit} designs a light-weight ViT combining CNNs for mobile devices. 
  Note that we employ our ensemble method to aggregate original MobileViT-S for fair comparison, denoted as MobileViT-S-Ens. 
\end{itemize}

\begin{table*}
  \centering
  \renewcommand{\arraystretch}{0.9}
  \setlength{\extrarowheight}{0pt}
  \addtolength{\extrarowheight}{\aboverulesep}
  \addtolength{\extrarowheight}{\belowrulesep}
  \setlength{\aboverulesep}{0pt}
  \setlength{\belowrulesep}{0pt}
  \caption{Comparison of On-device Evaluation Between Our Methods and Large ViTs Using ImageNet-1K Datasets}
  \vspace{-5pt}
  \label{tab:device_large_im1k}
  \begin{tabular}{c|cccccc} 
  \toprule
  \textbf{Model}                                     & \textbf{\#Params.} & \textbf{FLOPs} & \textbf{Accuracy} $\uparrow$ & \textbf{Latency} $\downarrow$         & \textbf{Energy} $\downarrow$            & \textbf{Power} $\downarrow$          \\ 
  \hline
  ViT-L/16                                           & 304.0 M            & 190.7 G        & 76.50 \%          & -                         & -                           & -                        \\
  \rowcolor[rgb]{0.949,0.949,0.949} \textbf{DeViTs}  & 23.0 M             & 5.9 G          & 74.85~\%          & \textbf{31.30}{\tiny$\pm$2.3} ms & \textbf{204.29}{\tiny$\pm$25.0} mJ & \textbf{6.46}{\tiny$\pm$0.6} W  \\ 
  \hline
  DeiT-B                                             & 86.6 M             & 35.1 G         & 83.40~\%          & -                         & -                           & -                        \\
  \rowcolor[rgb]{0.949,0.949,0.949} \textbf{DeDeiTs} & 23.9 M             & 5.9 G          & 81.84~\%          & \textbf{32.22}{\tiny$\pm$2.1} ms & \textbf{221.08}{\tiny$\pm$24.0} mJ & \textbf{6.86}{\tiny$\pm$0.7} W  \\ 
  \hline
  CCT-14\_7$\times$2                                & 22.4 M             & 11.1 G         & 80.67~\%          & 44.59{\tiny$\pm$4.9} ms          & 432.50{\tiny$\pm$51.8} mJ          & 9.69{\tiny$\pm$0.3} W           \\
  \rowcolor[rgb]{0.949,0.949,0.949} \textbf{DeCCTs}  & 18.3 M             & 5.1 G          & 79.12~\%          & \textbf{20.66}{\tiny$\pm$1.6} ms & \textbf{165.09}{\tiny$\pm$17.5} mJ & \textbf{7.99}{\tiny$\pm$0.6} W  \\
  \bottomrule
  \end{tabular}
  \end{table*}

\begin{table*}
  \centering
  \renewcommand{\arraystretch}{0.9}
  \setlength{\extrarowheight}{0pt}
  \addtolength{\extrarowheight}{\aboverulesep}
  \addtolength{\extrarowheight}{\belowrulesep}
  \setlength{\aboverulesep}{0pt}
  \setlength{\belowrulesep}{0pt}
  \caption{Comparison of On-device Evaluation Between Our Methods and Lightweight Baseline Models Using ImageNet-1K Datasets}
  \vspace{-5pt}
  \label{tab:device_small_im1k}
  \begin{tabular}{c|cccccc} 
  \toprule
  \textbf{Model}                                     & \textbf{\#Params.} & \textbf{FLOPs} & \textbf{Accuracy} $\uparrow$ & \textbf{Latency} $\downarrow$ & \textbf{Energy} $\downarrow$   & \textbf{Power} $\downarrow$ \\ 
  \hline
  T2T-ViT$_t$-14                                     & 21.5~M             & 12.2 G         & 81.70~\%          & 44.66{\tiny$\pm$2.8} ms & 422.51{\tiny$\pm$33.8} mJ & 9.46{\tiny$\pm$0.4} W  \\
  Twins-SVT-S                                        & 24.0 M             & 2.9 G          & 81.70~\%          & 70.10{\tiny$\pm$5.5} ms & 661.71{\tiny$\pm$57.5} mJ & 9.44{\tiny$\pm$0.3} W  \\
  Twins-PCPCT-S                                      & 24.1~M             & 3.8 G          & 81.20~\%          & 66.77{\tiny$\pm$4.8} ms & 624.35{\tiny$\pm$53.6} mJ & 9.35{\tiny$\pm$0.3} W  \\
  MobileViT-S                                        & 5.0~M              & 2.0 G          & 78.40~\%          & 55.42{\tiny$\pm$5.6} ms & 494.41{\tiny$\pm$50.9} mJ & 8.92{\tiny$\pm$0.1} W  \\ 
  MobileViT-S-Ens                                    & 20.1~M             & 8.1 G          & 78.62~\%          & 57.84{\tiny$\pm$6.4} ms & 546.26{\tiny$\pm$53.5} mJ & 9.42{\tiny$\pm$0.2} W  \\ 
  \hline\hline
  \rowcolor[rgb]{0.949,0.949,0.949} \textbf{DeViTs}  & 23.0 M             & 5.9 G          & 74.85~\%          & 31.30{\tiny$\pm$2.3} ms & 204.29{\tiny$\pm$25.0} mJ & \textbf{6.46}{\tiny$\pm$0.6} W  \\
  \rowcolor[rgb]{0.949,0.949,0.949} \textbf{DeDeiTs} & 23.9 M             & 5.9 G          & \textbf{81.94~}\% & 32.22{\tiny$\pm$2.1} ms & 221.08{\tiny$\pm$24.0} mJ & 6.86{\tiny$\pm$0.7} W  \\
  \rowcolor[rgb]{0.949,0.949,0.949} \textbf{DeCCTs}  & 18.3 M             & 5.1 G          & 79.12~\%          & \textbf{20.66}{\tiny$\pm$1.6} ms & \textbf{165.09}{\tiny$\pm$17.5} mJ & 7.99{\tiny$\pm$0.6} W  \\
  \bottomrule
  \end{tabular}
  \vspace{-5pt}
  \end{table*}

\textbf{Evaluation metrics. }
We employ the Top-1 classification accuracy as the performance metric. 
The end-to-end latency, energy consumption and average power are utilized as the efficiency metrics. 
As the models are deployed, we use the PyTorch Profiler tool \cite{torch_profiler} to profile the end-to-end latency of one inference. 
The whole energy and power are measured by Monsoon High Voltage Power Monitor \cite{hvpm}, which can collect the output voltage and current at a sampling rate of 5,000 samples per second. 
Before measurement, we switch the NVIDIA Jetson Nano to the max power mode, i.e., the max power is 10 Watt, and turn off the dynamic voltage and frequency scaling to ensure a steady measurement environment. 
We perform inference on the test set of the ImageNet dataset using NVIDIA Jetson Nanos, and measure the inference overhead of 50,000 testing samples in 1,000 different categories. 
For each ViT model, we run it for once as a warm-up and then record the execution time with 100 runs without break for the whole testing set. 
The aim of warm-up running is to alleviate the impact of weight loading and PyTorch initiation since we have omitted these overheads. We minus the power of the standby system from the measured power to acquire the power of performing inference. 
Thus, the energy consumption of a single inference can be obtained by taking the product of the averaged latency and the power of performing inference.

\subsection{Evaluation on the GPU Server}
In this set of experiments, we measure the classification accuracy results on four widely-used datasets as well as end-to-end inference latency on the GPU server. 
We choose ViT-L/16, DeiT-B, and CCT-14\_7$\times$2 as large ViTs to be decomposed. 
These models are all decomposed into 4 small models. 
End-to-end latency is measured on a GPU server with NVIDIA RTX 3090 GPUs, which is the average value with 100 runs without break for the whole testing set of corresponding datasets. 
We present the results compared with ViTs of different model sizes as follows.

\textbf{Comparison with large ViTs. }
We compare our methods with the large ViTs for various ViT backbones in the GPU server. 
The results of accuracy and end-to-end latency on four datasets are shown in Table \ref{tab:result_gpu_large}. 
We can find that large ViTs achieve superior performance nevertheless with extremely high latency overheads and huge computation costs, 
while our methods can reduce computation costs and accelerate inference latency by 1.98 $\times$ on average with only a 2\% accuracy drop. 
In most real-world deployments, a 2\% accuracy loss is considered negligible given the considerable latency reduction. 
Specifically, DeViTs can run 2.89$\times$ faster with only 0.96\% accuracy sacrifice than ViT-L/16. 
Our methods can also accelerate end-to-end latency by approximately 1.99 $\times$ with a 1.55\% accuracy drop on average for different backbones even on the large-scale dataset, ImageNet-1K.
These results demonstrate that our framework can reduce the computation cost of large ViTs for different backbones.
Moreover, our models achieve real-time inference latency with only slight accuracy sacrifice on the GPU server.

\textbf{Comparison with lightweight ViTs. }
We compare our methods to a variety of representative baseline models with approximate model parameters. 
Table \ref{tab:result_gpu_small} presents the results of Top-1 classification accuracy and end-to-end inference latency on four datasets.
We can observe that the DeCCTs significantly outperforms all other lightweight ViTs with the lowest latency and acceptable accuracy. 
Specifically, DeViTs/DeDeiTs/DeCCTs can reduce inference latency by 34.47\%/31.49\%/42.77\%, respectively, compared with the state-of-the-art efficient ViT model, MobileViT-S, on ImageNet-1K, 
which shows that our methods achieve an excellent trade-off between efficiency and performance. 
On the other hand, we can also find that although the parameters or FLOPs of some models (\textit{i.e.} Twins-SVT-S, Twins-PCPCT-S, and MobileViT-S) are less than that of our models, 
their end-to-end latency results on the GPU server are not lower than our models. 
It suggests that the number of parameters or FLOPs is only indicative but cannot fully determine the inference efficiency, 
which may not be an appropriate criterion for inference latency.

\subsection{Evaluation on the Edge Device}

In this subsection, we evaluate our methods on the edge devices and measure the Top-1 classification accuracy, end-to-end inference latency, 
average energy consumption, and average power to compare our methods with large ViTs and other lightweight ViT models. 
We deploy these models on the NVIDIA Jetson Nanos and employ an external high voltage power monitor \cite{hvpm} for the measurement.
We first present the results of comparison with large ViTs and then compare our methods with lightweight ViTs. 

\textbf{Comparison with large ViTs. }
We compare our methods with the large ViTs for different ViT backbones on the edge device. 
Table \ref{tab:device_large_im1k} presents on-device comparison results on ImageNet-1K. 
ViT-L/16 and DeiT-B are infeasible for deployment on resource-constrained platforms like Jetson Nano due to their substantial memory requirements and computational costs, 
denoted by "-" in their latency, energy, and power results.
Despite the superior performance of larger ViTs, such as ViT-L/16 and DeiT-B, they remain nonviable for deployment on lower-performance edge devices. 
Our methods decompose the large ViTs for different backbones into multiple small models, facilitating efficient collaborative inference with accuracy nearly equivalent to the large ViTs.
Specifically, our DeCCTs model improves inference latency by 2.16$\times$ and decreases energy consumption by 61.83\% compared with CCT-14\_7$\times$2.
These results demonstrate that our methods can alleviate the deployment of large ViTs, significantly accelerating inference speed and reducing energy consumption with acceptable accuracy.

\textbf{Comparison with lightweight ViTs. }
We compare our methods with other representative lightweight ViTs on the edge device. 
Table \ref{tab:device_small_im1k} reports the on-device comparison results on ImageNet-1K. 
We can observe that our methods can achieve the lowest end-to-end inference latency and energy consumption with approximate accuracy while maintaining comparable accuracy with lightweight ViTs. 
For example, DeCCTs improves inference latency by 60.92\%, decreases energy consumption by 66.61\%, and improves accuracy by 0.72\% compared with the state-of-the-art lightweight ViT, MobileViT-S. 
Moreover, akin to the evaluation analysis on the GPU server, their end-to-end latency and energy consumption on real edge devices do not outperform our models despite the fewer number of parameters and FLOPs. 
Thus, there is no apparent correlation between the number of parameters or FLOPs and end-to-end latency and energy consumption of ViTs. 
The FLOPs and the number of parameters are not appropriate criteria to estimate the inference latency and energy consumption.

\begin{figure}[!t]
  \centering
  \includegraphics[width=0.35\textwidth]{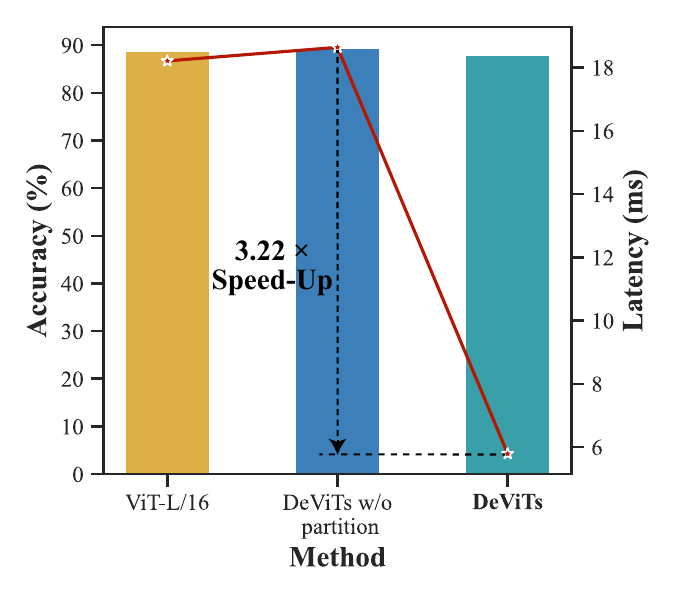}
  \vspace{-4pt}
  \caption{Ablation study of dataset partition on CIFAR-100. "DeViTs w/o partition" represents utilizing the original dataset.}
  \label{fig:ablation_partition}
\end{figure}

\subsection{Ablation Study}
\label{sec:ablation}

To fully understand the impact of each part of the proposed framework, we design ablation studies, where all experiments are evaluated on CIFAR-100 and utilize ViT as the backbone. 
We first evaluate the effectiveness of partitioning datasets to find the important parts of ViTs and study the impact of different loss functions in the sub-task distillation stage. 
Then we analyze the superiority of our ensemble methods by comparing different ensemble strategies and study the effect of different device numbers.

\textbf{Evaluation of dataset partition. }
We design an ablation experiment on CIFAR-100 to validate the effectiveness of utilizing the small partitioned datasets to find the important parts. 
The bar plot and line plot represent the results of accuracy and latency in Figure 2, respectively. ViT-L/16 is the backbone model and decomposed into four small models. 
We can find that utilizing the small datasets as the proxy to find the important parts can reduce redundant computation to accelerate inference by $3.22 \times$ with only a 1.73 \% accuracy drop compared to utilizing the original dataset. 
The results significantly validate the effectiveness of partitioning datasets and utilizing the small dataset as the proxy to find the important parts in decomposed models.

\begin{table}[!t]
  \centering
  \renewcommand{\arraystretch}{0.9}
  \setlength{\extrarowheight}{0pt}
  \addtolength{\extrarowheight}{\aboverulesep}
  \addtolength{\extrarowheight}{\belowrulesep}
  \setlength{\aboverulesep}{0pt}
  \setlength{\belowrulesep}{0pt}
  \caption{Ablation Study of Different Loss Functions in Sub-task Distillation}
  \vspace{-5pt}
  \label{tab:ablation_sub-task}
  \begin{tabular}{c|cccc} 
  \toprule
  \textbf{Model}                                                                & \textbf{GT}                                  & $\mathcal{L}_{pred}$                               & $\mathcal{L}_{feat}$                               & \textbf{Average Acc.}                                  \\ 
  \hline
  \begin{tabular}[c]{@{}c@{}}ViT-L/16 (Teacher)\end{tabular}                   & $\checkmark$                                 & $\times$                                         & $\times$                                         & 92.92 \%                                               \\ 
  \hline
  \begin{tabular}[c]{@{}c@{}}DeViT-Sub w/o\\distillation\end{tabular}           & $\checkmark$                                 & $\times$                                         & $\times$                                         & 85.92 \%                                               \\ 
  \hline
  \multirow{3}{*}{\begin{tabular}[c]{@{}c@{}}DeViT-Sub\\(Student)\end{tabular}} & $\times$                                     & $\checkmark$                                     & $\times$                                         & 87.20 \%                                               \\
                                                                                & $\times$                                     & $\times$                                         & $\checkmark$                                     & 89.57 \%                                               \\
                                                                                & {\cellcolor[rgb]{0.949,0.949,0.949}}$\times$ & {\cellcolor[rgb]{0.949,0.949,0.949}}$\checkmark$ & {\cellcolor[rgb]{0.949,0.949,0.949}}$\checkmark$ & {\cellcolor[rgb]{0.949,0.949,0.949}}\textbf{90.15 \%}  \\
  \bottomrule
  \end{tabular}
  \vspace{-5pt}
  \end{table}

\begin{table}[!t]
  \centering
  \renewcommand{\arraystretch}{0.9}
  \setlength{\extrarowheight}{0pt}
  \addtolength{\extrarowheight}{\aboverulesep}
  \addtolength{\extrarowheight}{\belowrulesep}
  \setlength{\aboverulesep}{0pt}
  \setlength{\belowrulesep}{0pt}
  \caption{Ablation Study of Different Loss Functions in Model Ensemble}
  \vspace{-5pt}
  \label{tab:ablation_ensemble}
  \begin{tabular}{c|cccc} 
  \toprule
  \textbf{Model}                                                  & \textbf{GT}                                  & $\mathcal{L}_{pred}$                               & $\mathcal{L}_{token}$                              & \textbf{Accuracy}                                      \\ 
  \hline
  ViT-L/16                                                        & $\checkmark$                                 & $\times$                                         & $\times$                                         & 88.77 \%                                               \\ 
  \hline
  \begin{tabular}[c]{@{}c@{}}DeViTs w/o\\supervision\end{tabular} & $\checkmark$                                 & $\times$                                         & $\times$                                         & 85.76 \%                                               \\ 
  \hline
  \multirow{3}{*}{\textbf{DeViTs}}                                & $\times$                                     & $\checkmark$                                     & $\times$                                         & 86.03 \%                                               \\
                                                                  & $\times$                                     & $\times$                                         & $\checkmark$                                     & 86.46 \%                                               \\
                                                                  & {\cellcolor[rgb]{0.949,0.949,0.949}}$\times$ & {\cellcolor[rgb]{0.949,0.949,0.949}}$\checkmark$ & {\cellcolor[rgb]{0.949,0.949,0.949}}$\checkmark$ & {\cellcolor[rgb]{0.949,0.949,0.949}}\textbf{87.98 \%}  \\
  \bottomrule
  \end{tabular}
  \vspace{-5pt}
  \end{table}

\textbf{Evaluation of sub-task distillation. }
The ablation study results of sub-task distillation are shown in Table \ref{tab:ablation_sub-task}. 
In this part, we choose the ViT-L/16 trained on different small partitioned datasets as the teacher for the corresponding small models. 
The backbone of each model is ViT. 
In the table, DeViT represents the decomposed small models for the backbone of ViT. 
GT means that the ground truth of the dataset is used as labels and distillation is not utilized. 
The Top-1 accuracy is the average results in the test set of all small partitioned dataset. 
From the results, we find that the small and shallow models can not learn deep and rich knowledge by themselves. 
With the help of large and powerful models, small models as students can learn the rich information effectively and acquire large performance improvement by distillation. 
Even if we only use predictions of the teacher for supervision, the average Top-1 accuracy can be enhanced by 1.28\%. 
The results of using $\mathcal{L}_{feat}$ indicates that distillation based on intermediate features can promote small models to learn more fine-grained knowledge and improve classification performance. 
These two loss function $\mathcal{L}_{pred}$ and $\mathcal{L}_{feat}$ are both helpful, and the best result is achieved when both terms are adopted.
The improvement is 4.23\% compared to training without distillation.

\textbf{Evaluation of model ensemble. }
In this part, we evaluate the performance of the model ensemble stage. 
Ablation study results under the supervision of powerful teacher models are shown in Table \ref{tab:ablation_ensemble}. 
ViT-L/16 is a large powerful ViT and decomposed into four small models.
It is also the teacher that guides the ensemble of all small decomposed models. 
Without the help of distillation, the direct ensemble of ViTs can lead to large accuracy loss. 
Compared to ViT-L/16, the accuracy of the proposed method without distillation decreases by 3.01\%. 
The results show that distillation can improve the ensemble performance of small models and reduce the accuracy loss caused by decomposition and shrinking. 
We use two types of loss functions to help the ensemble learning of small decomposed models. 
Each loss function is helpful, and the best result is achieved when both terms are adopted.
The minimal accuracy loss is 0.79\% when both two loss functions are adopted. 

The comparison results of different ensemble strategies are shown in Table \ref{tab:compare_ensemble}. 
Three competitive ensemble strategies are compared: 
(i) The \textit{Attention} method takes a simple key-value form to aggregate features. 
(ii) The \textit{SENet} \cite{senet} computes the importance of different features and weightedly sum these features. 
(iii) The \textit{MLP} approach uses an MLP block to fuse different features. 
From the table, we observe that SENet achieves the highest accuracy but also the highest inference latency compared to MLP. 
We remove the activation function in the MLP block and find that it can improve latency by 1.06$\times$ with only 0.03\% accuracy loss. 
The slight accuracy loss is negligible compared with the greatly reduced latency. 
This demonstrates that our method can effectively trade off latency and accuracy.

\begin{figure}[!t]
    \centering
    \includegraphics[width=0.4\textwidth]{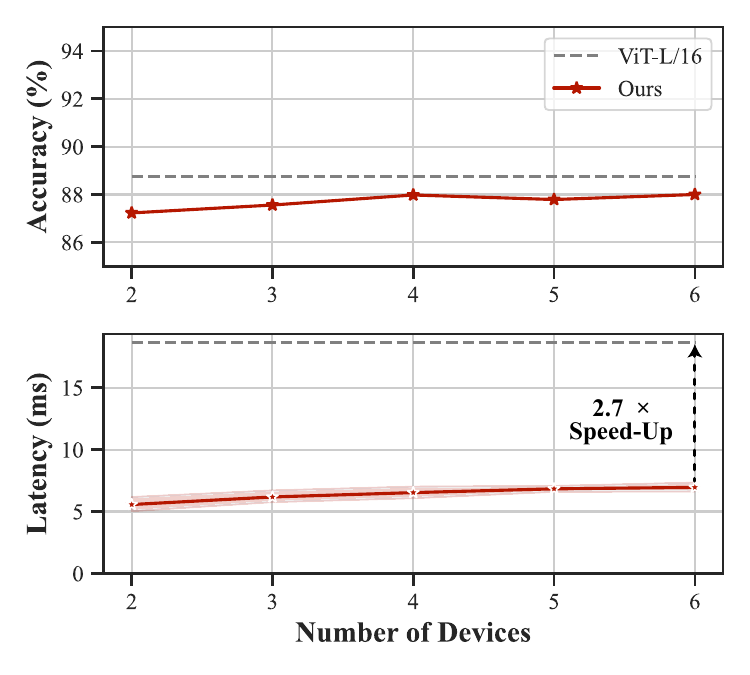}
    \vspace{-4pt}
    \caption{Results of accuracy and latency for different number of devices. The size of decomposed small models is the same.}
    \label{fig:device_num}
  \end{figure}

\textbf{Effect of device number}. We analyze the influence on accuracy and latency when changing the number of devices. 
We employ ViT-L/16 as the decomposed huge ViT and measure the classification accuracy and inference latency in the GPU server varying the number of devices from 2 to 6. 
The results on CIFAR-100 is shown in Figure \ref{fig:device_num}. As the number of devices increases, the accuracy is closer to the huge ViT while the latency remains essentially consistent. 
Our method can achieve $2.7 \times$ speed-up with only an approximate 1 \% accuracy drop even if there are 6 devices. The result verifies that our method is robust to the number of device number and achieve efficient inference.

\begin{table}[!t]
  \centering
  \renewcommand{\arraystretch}{0.9}
  \setlength{\extrarowheight}{0pt}
  \addtolength{\extrarowheight}{\aboverulesep}
  \addtolength{\extrarowheight}{\belowrulesep}
  \setlength{\aboverulesep}{0pt}
  \setlength{\belowrulesep}{0pt}
  \caption{Comparison of Different Ensemble Methods}
  \vspace{-5pt}
  \label{tab:compare_ensemble}
  \begin{tabular}{c|cc} 
  \toprule
  \textbf{Ensemble Methods}                       & \textbf{Accuracy} & \textbf{Average Latency}        \\ 
  \hline
  Attention                              & 87.09 \% & 6.77{\tiny $\pm$0.4} ms           \\ 
  SENet                                  & 88.05 \% & 6.17{\tiny $\pm$0.6} ms           \\ 
  MLP                                    & 88.01 \% & 6.16{\tiny $\pm$0.4} ms           \\ 
  \rowcolor[rgb]{0.949,0.949,0.949} \textbf{Ours} & 87.98 \% & \textbf{5.79}{\tiny $\pm$0.1} ms  \\
  \bottomrule
  \end{tabular}
  \vspace{-5pt}
  \end{table}

\section{Related Works}
\label{sec:related_works}
\subsection{Efficient ViT}
ViTs achieve superior performance by substantially increasing the number of parameters in intelligent applications \cite{DBLP:conf/sc/NarayananSCLPKV21}. 
We analyze the prior work of ViT training on a multi-GPU server and efficient ViT inference. Then we compare these methods with ours. 

\textbf{ViT training. }
Training large-scale ViTs is challenging due to the limitations imposed by GPU memory capacity and substantial computational cost. 
Most studies for training these transformers within a multi-GPU server aim to reduce GPU memory requirements in a single GPU and communication consumption, hence promoting efficient training.  
These approaches leverage parallelism strategies and fall into three broad categories: 
\textit{i) Data parallelism. }
Shoeybi \textit{et al.} \cite{DBLP:journals/corr/abs-1909-08053} proposed to partition the original dataset into small subsets and distribute them across multiple GPUs for training.
\textit{ii) Model parallelism. }
The huge transformer model is split across and by neural network layers into partitions processed on separate GPUs \cite{DBLP:conf/cvpr/IandolaMAK16}, 
necessitating communication and transfer of intermediate results for inference or model training. 
\textit{iii) Pipeline parallelism. }
GPipe \cite{DBLP:conf/nips/HuangCBFCCLNLWC19} utilizes pipeline parallelism by dividing consecutive groups of layers into cells positioned on separate GPUs.

\textbf{ViT inference. }
Given the impressive performance of large-scale ViTs, there is an urgent need to compress them for efficient edge inference. 
Recent studies have proposed lightweight architectures to enhance performance. 
For instance, Mehta \textit{et al.} \cite{mobile_vit} incorporated convolution into transformers, combining the strengths of convolution and attention. 
Moreover, a series of methods apply traditional model compression techniques to achieve compact ViTs, such as network pruning \cite{vtp}, knowledge distillation \cite{DBLP:conf/nips/HaoGJ0T00022, DBLP:journals/tmm/HaoLWHA22} and low-bit quantization \cite{DBLP:conf/nips/YaoAZWLH22}.  
Zhu \textit{et al.} \cite{vtp} introduced learnable coefficients to evaluate the importance of linear matrix, and neurons with small coefficients are removed.
Hao \textit{et al.} \cite{DBLP:conf/nips/HaoGJ0T00022} utilized patch-level information to help compact student models imitate teacher models. 
Yao \textit{et al.} \cite{DBLP:conf/nips/YaoAZWLH22} proposed an efficient and affordable post-training quantization approach to compress large transformers. 

In terms of ViT training on a multi-GPU server, 
the effectiveness of data parallelism is constrained by network bandwidth and the size of model. 
While model parallelism and pipeline parallelism can reduce the size of the model on a single GPU, the communication overhead remains substantial due to frequent data transmission between different nodes.
Regrading to efficient ViT inference, these methods omit the divisibility of ViTs and cannot achieve a satisfactory trade-off between performance and efficiency. 
To mitigate these limitations, we propose to decompose a large-scale ViT into multiple smaller models, each tailored based on the significance of their structural components. 
These compact models are executed in parallel and their intermediate results are transferred to the central node for only a single communication. 
Moreover, we first develop a novel feature matching module that facilitates the learning of decomposed models from the large-scale ViT, thereby minimizing information loss.

\subsection{Collaborative Inference}
Collaborative inference involves partitioning a DNN into two or more segments and deploying them across various devices for inference. 
This partitioning of DNNs can be executed layer-wise or across layers. 
We investigate these two types of approaches, layer-wise splitting and across layer splitting, followed by a comparison with our methods. 

\textbf{Layer-wise splitting.} 
A DNN is divided into multiple layer groups and distributed across different mobile edge nodes. 
The front part of the DNN is first executed on the lower-performance edge device, and the obtained intermediate features are offloaded to an edge server or the cloud. 
Most work in this category focuses on deploying partitioned DNNs into distinct devices based on their computational performance, aiming for an optimal balance between performance and efficiency \cite{Neurosurgeon, DBLP:conf/sensys/Yao0LWLSA20, DBLP:journals/tmc/TangW22,cnnpc}. 
For example, Kang \textit{et al.} \cite{Neurosurgeon} demonstrated that judicious selection of a splitting point can decrease latency and energy consumption compared to executing DNNs entirely on the cloud or mobile devices.
Tang \textit{et al.} \cite{cnnpc} proposed a deep reinforcement learning algorithm to obtain the optimal task offloading strategy. 
Another line of work employs quantization \cite{DBLP:conf/icpads/LiHJWWZ18} or encoding methods \cite{DBLP:conf/sensys/Yao0LWLSA20} to compress intermediate features. 
Yao \textit{et al.} \cite{DBLP:conf/sensys/Yao0LWLSA20} utilized an auto-encoder to compress features and then decode them in the cloud. 
Hao \textit{et al.} \cite{hao2022multi} combined quantization and auto-encoder to facilitate communication overheads, and designed an MAHPPO algorithm to solve the multi-agent collaborative inference problem. 

\textbf{Across layer splitting.}
The DNN weight is generally split into a hierarchy of multiple groups that have no connection among them, allowing parallel execution and model size reduction. 
These methods require an additional training procedure where publicly available pre-trained models cannot be used after splitting, compared to layer-wise splitting. 
For instance, Kim \textit{et al.} \cite{DBLP:conf/icml/KimPKH17} split the DNNs into either a set or a hierarchy of multiple groups to create a tree-structured model. 
These partitioned models can be executed in parallel to accelerate inference. 
Some methods utilize progressive slicing mechanisms to partition neural models into multiple components to fit in heterogeneous devices. 
Kim \textit{et al.} \cite{DBLP:conf/eurosys/KimKCKK19} proposed an inference runtime by splitting DNNs across layers. 
The partitioned models are deployed on the heterogeneous processors of an edge device. 
Mohammed \textit{et al.} \cite{DBLP:conf/infocom/MohammedJBF20} divided a DNN into multiple partitions that can be processed locally by end devices or offloaded to one or multiple powerful nodes. 

While layer groups split by layers are easily deployed on edge devices, layer splits necessitate the transfer of semi-processed input to the subsequent node and the output to the user, thereby increasing the overall execution time.  
Across layer splitting methods require frequent intercommunication among network splits, considerably increasing the communication overheads.
Furthermore, these methods primarily focus on CNNs and cannot be directly applied to the deployment of ViTs. 
To the best of our knowledge, we first propose a collaborative inference framework for general ViTs in edge devices by decomposing the large ViT into multiple small models.

\section{Conclusion}
\label{sec:conclusion}

In this paper, we proposed a novel framework to achieve efficient collaborative inference for ViTs, 
where we first decompose the large ViT into multiple small models and deploy these models on edge devices. 
Then we designed the DEKD ensemble algorithm to fuse multiple decomposed models and compensate for the accuracy loss caused by decomposition. 
We also designed a feature matching module to facilitate intermediate knowledge distillation given heterogeneous backbones. 
We conducted extensive experiments to verify the effectiveness of the proposed framework. 
The results showed that our framework can dramatically accelerate the inference and reduce the energy consumption on edge devices, 
which is difficult for existing works. 
Considering that we only focus on the homogeneous devices in this work, 
we intend to analyze the effect of device heterogeneity and explore how to achieve efficient collaborative inference on heterogeneous edge devices in the future.



\ifCLASSOPTIONcaptionsoff
  \newpage
\fi

\bibliographystyle{IEEEtran}
\bibliography{references.bib}

\clearpage

\appendices
\section{Preliminaries of ViT}
\label{sc:vit_intro}
The ViT model consists of a patch embedding layer, multiple stacked transformer encoders, 
and a classification head, as shown in Fig. \ref{fig:vit}. 
For image classification, a standard ViT first splits an image into fixed-size patches. 
Then these patches are flattened and mapped to $D$-dimensional feature vectors using a trainable linear projection or CNN blocks\cite{vit,deit}, known as the patch embeddings. 
A learnable embedding, termed as $[class]$ token, 
is appended to the sequence of patch embeddings to serve as the whole image representation.
Then, learnable 1D position embeddings are added to the patch embeddings to retain positional information.
The combined sequence of embedding vectors are fed into the subsequent transformer encoders. 
Finally, The classification head implemented by a single linear layer is used to gain the final classification result.  

The transformer encoder\cite{transformer} is the most important module in ViT, 
which is composed of multiple alternating layers of multiheaded self-attention (MSA) and multi-layer perceptron (MLP) blocks. 
The input sequence to the encoder is first passed to LayerNorm (LN), and then passed to the MSA and MLP blocks. 
The residual connections are used after every block. 
The output of the encoder is passed to another encoders until the final results are obtained. 

\textbf{Multiheaded Self-Attention (MSA).} 
Multiheaded self-attention is an extension of self-attention. 
In the standard self-attention, given the input sequence $\mathbf{Z} \in \mathbb{R}^{M\times D} $, 
the query vector $\mathbf{B}  \in \mathbb{R}^{M\times D_h}$, 
key vector $\mathbf{K} \in \mathbb{R}^{M\times D_h}$, 
and value vector $\mathbf{V} \in \mathbb{R}^{M\times D_h}$ are generated by applying linear transformation to the input sequence $\mathbf{Z}$, 
where $M$ is the number of patches, $D$ is the embedding dimension, and $D_h$ is the dimension of these vectors. 
Then we compute scores $\mathbf{s}$ between query and key vectors as
\begin{equation}
  \mathbf{s}=\frac{\mathbf{Q}\mathbf{K}^T}{\sqrt[]{d_h}} ,
\end{equation}
where $\frac{1}{\sqrt[]{d_h}}$ is the scaling factor. 
The scores are transformed into probabilities using softmax function, i.e., $\mathbf{S}=\text{softmax}\left(\mathbf{s}\right)$. 
Finally, the output of self-attention can be obtained by computing a weighted sum over value vectors with $\mathbf{S}\mathbf{V}$. 
The process can be unified into an equation: 
\begin{equation}
  \text{Self-Attention}(\mathbf{Q},\mathbf{K},\mathbf{V})=\text{softmax}\left(\frac{\mathbf{Q}\mathbf{K}^T}{\sqrt[]{d_h}}\right) \mathbf{V}.
\end{equation}
In MSA, $h$ is the number of heads. 
On each of the queries, keys and values, we perform $h$ self-attention operations in parallel, 
and apply projection to their concatenated outputs: 

\begin{equation}
  \begin{aligned}
    \text{MultiHead}\left(\mathbf{Q'},\mathbf{K'},\mathbf{V'}\right)&=\text{Concat}\left(\text{head}_1,\ldots,\text{head}_h\right)\mathbf{W}^o,\\
    \text{head}_i&=\text{Self-Attention}(\mathbf{Q}_i,\mathbf{K}_i,\mathbf{V}_i),
  \end{aligned}
\end{equation}
where the dimension of $\{\mathbf{Q}_i\}_{i=1}^{h},\{\mathbf{K}_i\}_{i=1}^{h}$, 
and $\{\mathbf{V}_i\}_{i=1}^{h}$ are $D_{q'}=D_{k'}=D_{v'}=D/h=64$. 
$\mathbf{Q'}$ is the concatenation of $\{\mathbf{Q'}_i\}_{i=1}^{h}$. 
$\mathbf{K'}$ and $\mathbf{V'}$ have the same form as $\mathbf{Q'}$. 
The dimension of the projection matrix $\mathbf{W}^o$ is $D\times D$.

\textbf{Multi-Layer Perceptron (MLP).} 
The MLP block consists of two fully-connected layers with a nonlinear activation function $\sigma(\cdot)$, 
usually being the GELU function. Let $\mathbf{Z}_{MLP}$ be the input of MLP.
The output of MLP can be expressed as
\begin{equation}
  \text{MLP}=\sigma\left(\mathbf{Z}_{MLP}\mathbf{W}_1+\mathbf{b}_1\right)\mathbf{W}_2+\mathbf{b}_2,
\end{equation} 
where $\mathbf{W}_1 \in \mathbb{R}^{D\times d}, \mathbf{b}_1\in \mathbb{R}^{d}, \mathbf{W}_2 \in \mathbb{R}^{d\times D}$, 
and $\mathbf{b}_2\in \mathbb{R}^{D}$ are the weights and biases of the first and second MLP layers, respectively.

\section{Additional Experiments}

\begin{table*}
  \centering
  \setlength{\extrarowheight}{0pt}
  \addtolength{\extrarowheight}{\aboverulesep}
  \addtolength{\extrarowheight}{\belowrulesep}
  \setlength{\aboverulesep}{0pt}
  \setlength{\belowrulesep}{0pt}
  \caption{Comparison of On-device Evaluation Between Our Methods and Large ViTs Using CIFAR-100 Datasets}
  \label{tab:device_large}
  \begin{tabular}{c|cccccc} 
  \toprule
  \textbf{Model}                                     & \textbf{\#Params.} $\downarrow$ & \textbf{FLOPs} $\downarrow$ & \textbf{Accuracy} $\uparrow$ & \textbf{Latency}  $\downarrow$                  & \textbf{Energy}   $\downarrow$                    & \textbf{Power}  $\downarrow$  \\ 
  \hline
  ViT-L/16                                           & 304.0 M            & 190.7~G        & 88.77~\%          & -                                   & -                                     & -                 \\
  \rowcolor[rgb]{0.949,0.949,0.949} \textbf{DeViTs}  & 23.0~M             & 5.9~G          & 87.98~\%          & 32.31{\tiny $\pm$6.0} ~ms                 & 234.92{\tiny $\pm$40.9}~mJ                  & 7.28{\tiny $\pm$0.2}~W  \\ 
  \hline
  DeiT-B                                             & 86.6~M             & 35.1~G         & 90.80~\%          & -                                   & -                                     & -                 \\
  \rowcolor[rgb]{0.949,0.949,0.949} \textbf{DeDeiTs} & 23.9~M             & 5.9~G          & 89.34~\%          & 35.25{\tiny $\pm$8.2}~ms                  & 274.82{\tiny $\pm$71.0}~mJ                  & 7.77{\tiny $\pm$0.5}~W  \\ 
  \hline
  CCT-14\_7$\times$ 2                                & 22.4~M             & 11.1~G         & 88.78~\%          & 44.66{\tiny $\pm$7.3}~ms                  & 379.88{\tiny $\pm$65.8}~mJ                   & 9.50{\tiny $\pm$0.5}~W   \\
  \rowcolor[rgb]{0.949,0.949,0.949} \textbf{DeCCTs}  & 18.3~M             & 5.1~G          & 87.43~\%          & \textbf{\textbf{23.54}}{\tiny $\pm$5.1}~ms & \textbf{\textbf{203.06}}{\tiny $\pm$47.5}~mJ & 8.61{\tiny $\pm$0.5}~W   \\
  \bottomrule
  \end{tabular}
  \end{table*}

\begin{table*}
  \centering
  \setlength{\extrarowheight}{0pt}
  \addtolength{\extrarowheight}{\aboverulesep}
  \addtolength{\extrarowheight}{\belowrulesep}
  \setlength{\aboverulesep}{0pt}
  \setlength{\belowrulesep}{0pt}
  \caption{Comparison of On-device Evaluation Between Our Methods and Lightweight Baseline Models Using CIFAR-100 Datasets}
  \label{tab:device_small}
  \begin{tabular}{c|cccccc} 
  \toprule
  \textbf{Model}                                     & \textbf{\#Params.} $\downarrow$ & \textbf{FLOPs} $\downarrow$ & \textbf{Accuracy} $\uparrow$ & \textbf{Latency}  $\downarrow$        & \textbf{Energy}  $\downarrow$           & \textbf{Power}  $\downarrow$         \\ 
  \hline
  T2T-ViT$_t$-14                                     & 21.5~M              & 12.2~G         & 86.62~\%          & 49.53{\tiny $\pm$8.4} ms          & 460.68{\tiny $\pm$78.5}~mJ          & 9.30{\tiny $\pm$0.1}~W           \\
  Twins-SVT-S                                        & 24.0~M              & 2.9~G          & 86.72~\%          & 85.66{\tiny $\pm$14.6} ms         & 823.69{\tiny $\pm$141.0} mJ         & 9.61{\tiny $\pm$0.2} W           \\
  Twins-PCPCT-S                                      & 24.1~M              & 3.8~G          & 86.70~\%          & 73.80{\tiny $\pm$12.1} ms         & 664.75{\tiny $\pm$109.7}~mJ         & 9.01{\tiny $\pm$0.1}~W           \\
  MobileViT-S                                        & 5.0~M               & 2.0~G          & 87.00~\%          & 61.14{\tiny $\pm$11.0} ms         & 606.13{\tiny $\pm$109.7}~mJ         & 9.91{\tiny $\pm$0.1} W           \\
  MobileViT-S-Ens                                    & 20.1~M              & 8.1~G          & 85.80~\%          & 62.25{\tiny $\pm$11.1} ms         & 638.06{\tiny $\pm$110.4} mJ         & 10.25{\tiny $\pm$0.3}~W          \\ 
  \hline\hline
  \rowcolor[rgb]{0.949,0.949,0.949} \textbf{DeViTs}  & 23.0~M              & 5.9~G          & 87.98~\%          & 32.31{\tiny $\pm$6.0} ms          & 234.92{\tiny $\pm$40.9} mJ          & \textbf{7.28}{\tiny $\pm$0.2}~W  \\
  \rowcolor[rgb]{0.949,0.949,0.949} \textbf{DeDeiTs} & 23.9~M              & 5.9~G          & \textbf{89.34~}\% & 35.25{\tiny $\pm$8.2} ms          & 274.82{\tiny $\pm$71.0} mJ          & 7.77{\tiny $\pm$0.5} W           \\
  \rowcolor[rgb]{0.949,0.949,0.949} \textbf{DeCCTs}  & 18.3~M              & 5.1~G          & 87.43~\%          & \textbf{23.54}{\tiny $\pm$5.1}~ms & \textbf{203.06}{\tiny $\pm$47.5} mJ & 8.61{\tiny $\pm$0.5} W           \\
  \bottomrule
  \end{tabular}
  \end{table*}

\subsection{Training settings}
The training settings vary in different backbones and datasets. 
We list the different hyperparameters in Table \ref{hyper}, 
and introduce others as follows. 
The models in this work are all firstly pretrained on ImageNeT-1K\cite{imagenet_1k}. 
During training, each sample is first resized to 256$\times$256, 
and then use Rand-Augment and random erasing are applied for data augmentation. 
The models are all trained using AdamW optimizer. 
We use the cosine scheduler to adjust the learning rate. 
The minimum learning rate is set as $1\times10^{-5}$. 
All the compared approaches are run with timm \cite{timm}, a GitHub repository, for fair comparison.

\begin{table}
    \centering
    \renewcommand{\arraystretch}{1.2}
    \caption{The Training Settings for Sub-distillation and Model Ensemble}
    \label{hyper}
    \begin{tabular}{c|c|ccc} 
    \toprule
    \multirow{2}{*}{Stage}            & \multirow{2}{*}{Hyperparameters} & \multicolumn{3}{c}{Backbones}  \\ 
    \cline{3-5}
                                      &                                  & ViT  & DeiT & CCT              \\ 
    \hline
    \multirow{5}{*}{Sub-distillation} & learning rate                    & 3e-4 & 5e-6 & 3e-3             \\
                                      & batch size                       & 128  & 128  & 256              \\
                                      & weight decay                     & 0    & 0.05 & 3e-2             \\
                                      & epochs                           & 500  & 1000 & 200              \\
                                      & max gradient                     & 1    & 0    & 0                \\ 
    \hline
    \multirow{5}{*}{Model Ensemble}   & learning rate                    & 6e-4 & 3e-4 & 8e-3             \\
                                      & batch size                       & 32   & 128  & 128              \\
                                      & weight decay                     & 0    & 0    & 3e-2             \\
                                      & epochs                           & 500  & 500  & 300              \\
                                      & max gradient                     & 1    & 1    & 0                \\
    \bottomrule
    \end{tabular}
    \end{table}

\subsection{On-device Evaluation on CIFAR-100}

\textbf{Comparison with large ViTs. }
The on-device comparison results between our models and large ViTs are shown in Table \ref{tab:device_large}. 
ViT-L/16 and DeiT-B cannot be deployed for inference on the resource-constrained Jetson Nano because of their high computation costs. 
Hence, we use "-" to represent the results of latency, energy, and power of ViT-L/16 and DeiT-B. 
Our methods can decompose large models to promote their efficient deployment with only 1.2\% accuracy loss on average. 
Compared with CCT-14\_7$\times$2, our model DeCCTs can improve inference latency by 1.9$\times$ and reduces energy consumption by 26.11\%. 
These results demonstrate that our methods can significantly accelerate inference speed and reduce energy consumption with an acceptable accuracy. 

\textbf{Comparison to lightweight ViTs. }
Table \ref{tab:device_small} reports the on-device comparison results between our models and the state-of-the-art lightweight ViTs. 
It can be seen that our models can achieve the lowest inference latency and energy consumption while the highest accuracy compared with the counterparts. 
Compared with T2T-ViT$_t$-14, our model DeCCTs can improve inference latency by 52.47\%, reduce energy consumption by 55.92\%, and improve accuracy by 0.81\%. 
Similar to the evaluation on the GPU server, 
although the number of parameters and FLOPs of MobileViT-S are both smaller than those of our models, the latency of MobileViT-S is not lower than our models. 
The result also indicates that the parameters and FLOPs are not appropriate to justify the inference latency and energy consumption.

\subsection{Comparison with Pruning Methods}
We compare our shrinking methods with other pruning methods for transformers to validate the effectiveness. The results of classification accuracy and throughput on ImageNet-1K are shown in Table \ref{tab:ablation_prune}. 
We employ the DeiT-B as the backbone ViT of all methods for fair comparison. 
We can see that our method can achieve higher classification accuracy and faster throughput with less computation complexity compared to other pruning methods.

\begin{table}
    \centering
    \setlength{\extrarowheight}{0pt}
    \addtolength{\extrarowheight}{\aboverulesep}
    \addtolength{\extrarowheight}{\belowrulesep}
    \setlength{\aboverulesep}{0pt}
    \setlength{\belowrulesep}{0pt}
    \caption{Comparison Between Ours and Other Shrinking Methods in ImageNet-1K}
    \label{tab:ablation_prune}
    \begin{tabular}{c|cccc} 
    \toprule
    \textbf{Method}                                 & \textbf{FLOPs} $\downarrow$ & \textbf{Top-1 Acc.} $\uparrow$ & \textbf{Top-5 Acc.} $\uparrow$ & \textbf{Throughput} $\uparrow$ \\ 
    \hline
    DeiT-B                                          & 17.6 G         & 81.8 \%             & 95.6 \%             & 292 img/s            \\
    \hline
    PoWER \cite{DBLP:conf/icml/GoyalCRCSV20}                                          & 10.4 G         & 80.1 \%             & 94.6 \%             & 397 img/s            \\
    VTP \cite{vtp}                                            & 10.0 G         & 80.7 \%             & 95.0 \%             & 412 img/s            \\
    \rowcolor[rgb]{0.949,0.949,0.949} \textbf{Ours} & \textbf{9.9} G          & \textbf{80.8} \%              & \textbf{95.4} \%             & \textbf{707} img/s            \\
    \bottomrule
    \end{tabular}
    \end{table}

\vfill

\end{document}